%% file: ieee_paper.tex
\documentclass[a4paper,10pt,conference]{ieeeconf}

\usepackage{amsmath,amssymb,amsfonts}
\usepackage{graphicx}
\usepackage{todonotes}
\usepackage{mathtools}
\usepackage{pgfplots} 
\usepackage{multicol}
\usepackage{flushend}
\usepackage{tabularx}
\usepackage{multirow}
\usepackage{siunitx}
\usepackage[export]{adjustbox}
\usepackage{caption}
\usepackage{array, graphicx}

\usepackage[
    backend=biber,
    natbib=true,
    url=false, 
    doi=false,
    eprint=false,
    maxcitenames=1,
    sorting=none,
    isbn=false
]{biblatex}

\usepackage[
		a4paper,        
        left=0.75in,
        right=0.514in,
        top=1.458in,
        bottom=0.75in
]{geometry}

\newcommand\copyrighttext{%
	\scriptsize \textcolor{blue}{\textcopyright 2019 IEEE. Personal use of this material is permitted.  Permission from IEEE must be obtained for all other uses, in any current or future media, including reprinting/republishing this material for advertising or promotional purposes, creating new collective works, for resale or redistribution to servers or lists, or reuse of any copyrighted component of this work in other works}}
\newcommand\copyrightnotice{%
	\begin{tikzpicture}[remember picture,overlay]
	\node[anchor=north,yshift=-7.5pt] at (current page.north) {\fbox{\parbox{\dimexpr\textwidth-\fboxsep-\fboxrule\relax}{\copyrighttext}}};
	\end{tikzpicture}%
}

\graphicspath{ {figures/} }

\begin{document}

\IEEEoverridecommandlockouts %

\title{\LARGE \textbf{Lane-Merging Using Policy-based Reinforcement Learning and Post-Optimization}\\
}

\author{Patrick Hart$^{1}$, Leonard Rychly$^{1}$ and Alois Knoll$^{2}$%
	\thanks{$^{1}$Patrick Hart and Leonard Rychly are with the fortiss GmbH, An-Institut Technische Universit\"{a}t M\"{u}nchen, Munich, Germany. Email: patrick.hart@tum.de, rychly@fortiss.org}%
	\thanks{$^{2}$Alois Knoll is with the Chair of Robotics, Artificial Intelligence and Real-time Systems, Technische Universit\"{a}t M\"{u}nchen, Munich, Germany}%
}

\maketitle
\copyrightnotice

\begin{abstract}
Many current behavior generation methods struggle to handle real-world traffic situations as they do not scale well with complexity. However, behaviors can be learned off-line using data-driven approaches. Especially, reinforcement learning is promising as it implicitly learns how to behave utilizing collected experiences. In this work, we combine policy-based reinforcement learning with local optimization to foster and synthesize the best of the two methodologies. The policy-based reinforcement learning algorithm provides an initial solution and guiding reference for the post-optimization. Therefore, the optimizer only has to compute a single homotopy class, e.g.\ drive behind or in front of the other vehicle. By storing the state-history during reinforcement learning, it can be used for constraint checking and the optimizer can account for interactions. The post-optimization additionally acts as a safety-layer and the novel method, thus, can be applied in safety-critical applications. We evaluate the proposed method using lane-change scenarios with a varying number of vehicles.
\end{abstract}

\section{Introduction}
As autonomous driving strives towards level-five autonomy novel behavior generation methods have to be developed to tackle the remaining challenges. One of these remaining challenges is to plan behaviors for complex traffic scenarios having multiple traffic participants.

Data-driven methods, such as policy-based reinforcement learning can learn how to behave in complex situations utilizing collected experiences. However, since the whole configuration space cannot be explored during training and most approaches use deep artificial neural networks, the reinforcement learning algorithm might fail to interpolate and to generalize well \cite{Boyan}. This can lead to unwanted and unexpected behaviors making reinforcement learning infeasible to use in safety-critical applications. 

\begin{figure}[t]
\vspace{2mm}
\centering
\def\svgwidth{8.5cm}
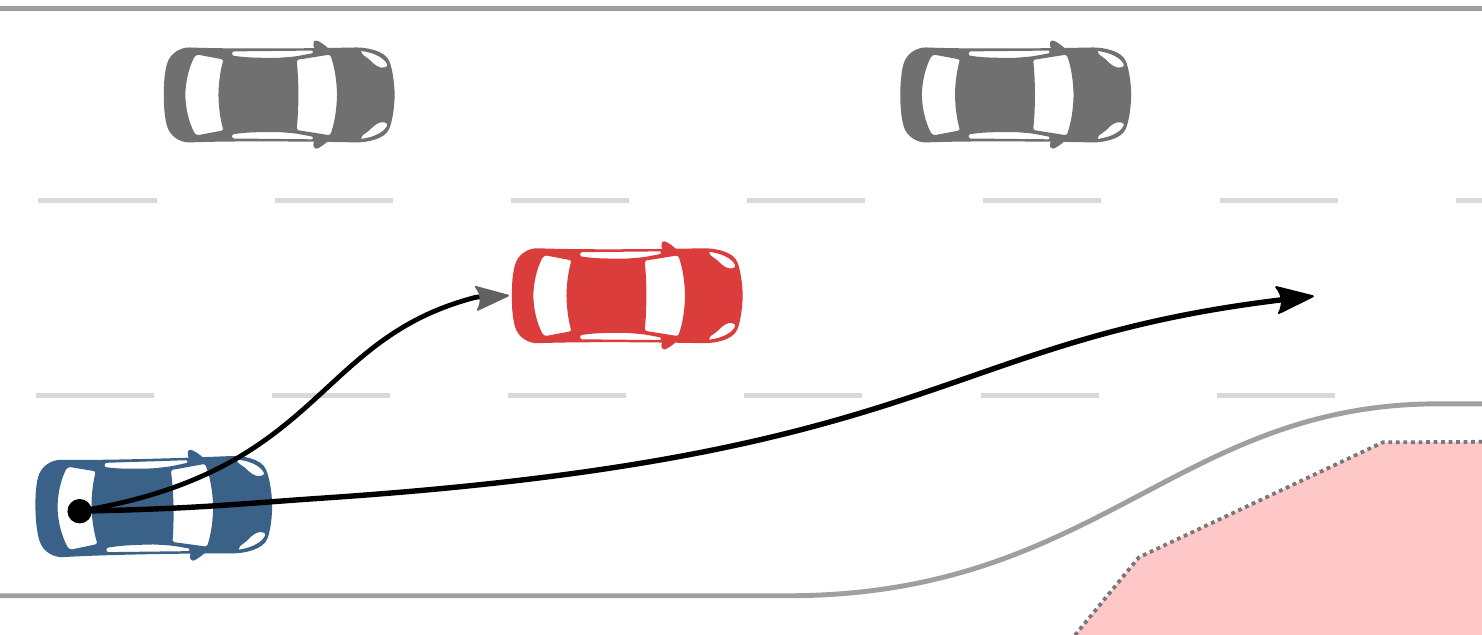
\caption{The blue vehicle wants to merge onto the highway. In this scenario, two homotopy classes exist -- merge in front of or behind the red vehicle.}
\label{fig:motivation}
\end{figure}

Contrary to this, conventional approaches, such as optimization and search-based techniques provide optimal and safe solutions given the problem is well defined. However, these approaches might fail to handle complex situations in real-time since they do not scale well with the increasing complexity of the environment. A local non-linear optimizer would have to solve multiple homotopy classes to find the global optimum. Figure \ref{fig:motivation} shows a scenario where a local optimizer would have to evaluate at least two combinatorial options: merge behind or in front of the red vehicle. 

The main contributions of this work are the following: 
\begin{itemize}
\item Obtain a behavior for complex traffic situations using reinforcement learning, 
\item use this behavior to choose a homotopy class and to include interactions in the optimization and, 
\item make the reinforcement learning algorithm combined with the post-optimization feasible in safety-critical applications. 
\end{itemize}

In this work, we combine policy-based reinforcement learning with local non-linear optimization to synthesize the best of the two methodologies. The learned policy of the reinforcement learning algorithm is iteratively executed in the environment to generate an initial estimate and a guiding reference trajectory for the optimization. The state history of all vehicles is stored and later used for constraint checking in the optimization. Since all vehicles react to their surroundings according to their intrinsic behavior policy, the interactions between the traffic participants are implicitly included in the optimization. Additionally, due to the initial estimate proposed by the reinforcement learning algorithm, the optimizer only has to solve a single homotopy class -- making the optimization also feasible for very complex traffic situations. Moreover, the post-optimization acts as an additional safety-layer after the reinforcement learning algorithm. Therefore, a high level of safety and comfort is guaranteed and the novel approach is also applicable in safety-critical applications.

We demonstrate and evaluate our novel approach using interactive lane-merging scenarios having a varying number of vehicles. All vehicles in the simulation framework have a behavior policy, such as the Intelligent Driver Model \cite{RajamaniR2012, Treiber2000}. The ego-vehicle is controlled by the novel behavior generation method presented in this work. 

The work is further organized as follows: Section II gives a quick overview of state-of-the-art algorithms used in behavior generation with a special focus on reinforcement learning. Next, a problem definition for this work is provided. In Section IV, our combined reinforcement and optimization approach is introduced. Section V evaluates the performance of the novel method using a lane-merging scenario. Finally, a conclusion, discussion, and an outlook are provided in Section VI.

\section{Related Work}
This section gives a brief overview of state-of-the-art reinforcement learning methods -- with a special focus on behavior generation for autonomous driving. Moreover, a short overview of optimization-based techniques to generate behaviors for autonomous vehicles is given. 

Reinforcement learning can roughly be divided into two categories: policy- and value-based methods \cite{Bharath2017}. In this work, we focus on policy-based reinforcement learning since it achieves state-of-the-art performance in large configuration spaces and dynamic control problems \cite{Lillicrap2015, Abdolmaleki2018}. In the deterministic case, policy-based methods map states to actions and in the stochastic case, they create distributions over actions for each state.

The Trust Region Policy Optimization (TRPO) is a policy-based method that additionally restrains the next policy to be close to the previous one by using the Kullback-Leibler (KL) divergence \cite{Schulman2015}. Based on this the Proximal Policy Optimization (PPO) has been developed that also restricts the objective function, but by using a clipping function instead of using the KL divergence \cite{Schulman}. This is computationally more efficient than the TRPO and yields equally good or better results in a wide range of applications \cite{Klissarov2017, Heess}. 

However, the above-presented methods explore on-policy and are, thus, more likely to become stuck in local minima. Recently, policy-based methods have been suggested that explore off-policy and, therefore, mitigate this problem, such as the Maximum a Posteriori Policy Optimization (MPO) \cite{Abdolmaleki2018} and the Soft Actor-Critic (SAC) \cite{HaarnojaZAL18}. To leverage the off-policy exploration advantage, we use the SAC algorithm in this work.

Reinforcement learning has been successfully applied to generate behaviors for autonomous vehicles. \citet{Shalev-Shwartz2016} use deep reinforcement learning to find long term driving strategies. Other than using an option-graph, they showed that the long-term driving problem can be solved without assuming the Markov property. However, as most of the reinforcement learning methods use deep neural networks, they cannot be simply applied in safety-critical applications as the interpolation and generalization of these cannot be guaranteed \cite{Boyan}. 

Therefore, reinforcement learning has been combined with formal verification to ensure that only safe actions are executed by the reinforcement learning algorithm \cite{Mirchevska2018}. However, due to the curse of dimensionality in formal verification, these methods become untractable fast with a growing action and configuration space \cite{Correll}. 

Another approach trying to make reinforcement learning safe and applicable was introduced by \citet{Levine2013}. They use differential dynamic programming in order to generate suitable guiding samples to direct policy learning and to avoid poor local optima. However, with the characteristic of neural networks having a large number of local optima, unwanted behavior still can occur and generalization cannot be guaranteed \cite{Swirszcz2016}. 

Contrary to data-driven reinforcement learning techniques, optimization-based behavior generation methods provide safe and optimal solutions. These are, therefore, already used in safety-critical applications, such as autonomous driving. As shown by \citet{Bender2015} all combinatorial options (homotopy classes) have to be evaluated in order to find the global optimum when using local optimization. Therefore, these methods generally do not scale well with the increasing complexity of the environment and become infeasible for real-time applications. In this work, we combine reinforcement learning to propose an initial solution and, thus, also a homotopy class for the optimizer. Therefore, the optimizer only has to solve a single homotopy and local-optimization techniques become feasible in very complex traffic scenarios.

\section{Problem Statement}
The proposed behavior generation method consists of two main components:

\begin{enumerate}
\item An off-line stochastic policy-based reinforcement learning method and,
\item a local non-linear post-optimization.
\end{enumerate}
The stochastic policy-based reinforcement learning provides an approximately optimal solution for the Markov Decision Process (MDP). This is achieved by interacting with an environment, collecting experiences and improving its stochastic policy $\pi_\phi$ parametrized by an artificial neural network having the parameters $\phi$. This policy is then iteratively executed to generate a state-space trajectory $\underline{\mathrm{x}}^\mathrm{RL} = [\underline{\mathrm{x}}_0, \dots, \underline{\mathrm{x}}_\mathrm{N}]$ and a control input sequence $\underline{\mathrm{a}}^\mathrm{RL} = [\underline{\mathrm{a}}_0, \dots, \underline{\mathrm{a}}_\mathrm{N}]$ for the ego-vehicle having $\mathrm{N}$ time-steps. The other vehicles behave according to their defined policies $[\pi_0, \dots, \pi_\mathrm{M}]$ for $\mathrm{M}$ other vehicles. All agent's state-space trajectories $[\underline{\mathrm{x}}^\mathrm{other, 1}, \dots, \underline{\mathrm{x}}^\mathrm{other, M}]$ are stored as they are later used for constraint checking in the optimization.

The second part consists of the local non-linear post-optimization that uses the generated state-space trajectory $\underline{\mathrm{x}}^\mathrm{RL}$ and the control input sequence $\underline{\mathrm{a}}^\mathrm{RL}$ as a guiding reference and initial optimization vector, respectively. Due to this, a homotopy class is implicitly chosen for the optimizer and only a single homotopy has to be evaluated by the local-optimization method to find the globally optimal control input sequence $\underline{\mathrm{a}}^{*} = [\underline{\mathrm{a}}_0, \dots, \underline{\mathrm{a}}_\mathrm{N}]$ that produces the state-space trajectory $\underline{\mathrm{x}}^{*} = [\underline{\mathrm{x}}_0, \dots, \underline{\mathrm{x}}_\mathrm{N}]$. Constraints and interactive behavior with other traffic participants are integrated into the optimization using the recorded state-space trajectories $[\underline{\mathrm{x}}^\mathrm{other, 1}, \dots, \underline{\mathrm{x}}^\mathrm{other, M}]$.

\section{Our Method}
In this section, we introduce the used policy-based reinforcement learning and how we generate behaviors for autonomous vehicles. Moreover, we then go into detail of the post-optimization and how it is combined with the policy-based reinforcement learning algorithm.

\subsection{Policy-based Reinforcement Learning}
This section describes the stochastic policy-based reinforcement learning and how the environment's rewards are defined. In the reinforcement learning problem, we try to optimize the expected cumulative future reward for the stochastic policy denoted by $\pi_\phi$ parametrized by the neural network parameters $\phi$. The basic objective function expressing the expected cumulative future reward can be mathematically denoted as
\begin{equation}
\mathrm{R} = \sum_\mathrm{t} \mathbb{E}_{(\underline{\mathrm{s}}_t, \underline{\mathrm{a}}_t) \sim \pi_\phi}[r(\underline{\mathrm{s}}_t, \underline{\mathrm{a}}_t)]
\label{eq:cum_reward}
\end{equation}
with $r(\cdot)$ being the reward function, $\underline{\mathrm{s}}_t$ the concatenated state-space and $\underline{\mathrm{a}}_t$ the action of the ego-vehicle. However, in this work, we use an extended maximum entropy objective as introduced by \citet{HaarnojaZAL18}.

The rewards of the environment are designed to avoid crashes, provide a comfortable driving experience and to follow a given reference with a desired speed. Derived from these characteristics the reward function is defined as
\begin{equation}
\mathrm{\mathbf{R}}_\mathrm{cost} = \mathrm{\mathbf{R}}_\mathrm{ref} + \mathrm{\mathbf{R}}_\mathrm{comfort} + \mathrm{\mathbf{R}}_\mathrm{vel} + \mathrm{\mathbf{R}}_\mathrm{col}
\label{eq:reward_function}
\end{equation}
with $\mathrm{\mathbf{R}}_\mathrm{ref}$ being the deviation to the reference path, $\mathrm{\mathbf{R}}_\mathrm{comfort}$ rating the jerk of the trajectory, $\mathrm{\mathbf{R}}_\mathrm{vel}$ being the deviation to the desired velocity and $\mathrm{\mathbf{R}}_\mathrm{col}$ the penalty for collisions. All rewards besides the collision-term are modeled as
\begin{equation}
\mathrm{\mathbf{R}} = \mathrm{r}_\mathrm{max} - \Delta_\mathrm{desired}^2
\label{eq:reward}
\end{equation}
with $\Delta_\mathrm{desired}^2$ being the squared deviation to a desired value, such as the speed or reference path and $\mathrm{r}_\mathrm{max}$ being the maximum achievable reward. Equation \ref{eq:reward} will assume its maximum value if the deviation to the desired values is zero. The penalty for a collision $\mathrm{\mathbf{R}}_\mathrm{col}$ is selected to be a large negative scalar value.

The input state consists of the concatenated states of all vehicles $\underline{\mathrm{s}}_t = [\underline{\mathrm{x}}_0, \dots, \underline{\mathrm{x}}_\mathrm{M}]$ with $\mathrm{M}$ being the number of vehicles, $t$ the time-index and $\underline{\mathrm{x}}_i$ the state of the $i$-th vehicle. On the output layer of the policy network, we utilize a normal distribution $\pi_\phi(\underline{\mathrm{s}}_t) \sim \operatorname{N}(\cdot)$ whose mean is scaled to limit the output space. For scaling the mean to a defined range of minimum and maximum control values ($\mathrm{\underline{a}}_\mathrm{min}$ and $\mathrm{\underline{a}}_\mathrm{max}$), we use a hyperbolic tangent scaling function that can be denoted as

\begin{equation}
\mathrm{\underline{a}}_\mathrm{t} = \frac{(\mathrm{\underline{a}}_\mathrm{max} + \mathrm{\underline{a}}_\mathrm{min})}{2} +  \frac{(\mathrm{\underline{a}}_\mathrm{max} - \mathrm{\underline{a}}_\mathrm{min})}{2} \tanh{\pi_\phi(\underline{\mathrm{s}}_\mathrm{t})}.
\end{equation}

After training has been concluded, the policy $\pi_\phi(\underline{\mathrm{s}}_\mathrm{t})$ is greedily executed iteratively $\mathrm{N}$-times in order to create the state-space trajectory $\mathrm{\underline{x}}^\mathrm{RL}=[\underline{\mathrm{x}}_0, \dots, \underline{\mathrm{x}}_\mathrm{N}]$ and the control input sequence $\mathrm{\underline{a}}^\mathrm{RL} = [\underline{\mathrm{a}}_0, \dots, \underline{\mathrm{a}}_\mathrm{N-1}]$. During the greedy execution of the stoachastic policy $\pi_\phi$ the maximum values of the distribution are being used in each time-step maximizing Equation \ref{eq:cum_reward}. All used hyper-parameters, reward values and other characteristics are explained in detail in Section V.

\subsection{Non-linear Post-Optimization}
The non-linear optimization utilizes the behavior policy of the reinforcement learning algorithm to obtain an initial optimization vector and a guiding reference. With the assumption that reinforcement learning approximates an optimal solution for the Markov Decision Process (MDP), the optimized solution is enforced to be close to it. Thus, a homotopy class for the optimization is implicitly selected and also complex planning problems become feasible to be solved online and in real-time using local optimization. Additionally, the optimization checks constraints and smoothens the control inputs $\mathrm{\underline{a}}^\mathrm{RL}$ of the reinforcement learning.

The optimization problem can be mathematically denoted as
\begin{align}
\text{minimize } &\mathrm{J}(\mathrm{\underline{x}}) \\
\text{subject to } &\mathrm{\underline{\dot{x}}}(t) = F(\mathrm{\underline{x}}(t), \mathrm{\underline{a}}(t))
\end{align}

with $\mathrm{J}$ being the objective function and $F(\underline{x}(t), \underline{u}(t))$ the dynamic model of the ego-vehicle. The dynamic model $\mathrm{\underline{\dot{x}}}(t) = F(\mathrm{\underline{x}}(t), \mathrm{\underline{a}}(t))$ is forward integrated $\mathrm{N}$-times and a state-space trajectory $\mathrm{\underline{x}}^\mathrm{Opt.}$ is obtained having the same length as $\mathrm{\underline{x}}^\mathrm{RL}$. Thus, a dynamically feasible state-space trajectory is guaranteed. In this work, we use a simplified single track model as presented in \cite{Hart2018}. The function $\mathrm{J}(\mathrm{\underline{x}})$ is defined as
\begin{equation}
\mathrm{J}(\mathrm{\underline{x}}) = \mathrm{\mathbf{J}}_\mathrm{jerk} + \mathrm{\mathbf{J}}_\mathrm{col} + \sum_\mathrm{i = 0}^\mathrm{N} (\mathrm{\underline{x}}^\mathrm{Opt.}_\mathrm{i} - \mathrm{\underline{x}}^\mathrm{RL}_\mathrm{i})
\end{equation}
with $\mathrm{\mathbf{J}}_\mathrm{jerk}$ being the squared jerk cost of the optimized trajectory $\mathrm{\underline{x}}^\mathrm{Opt.}$, $\mathrm{\mathbf{J}}_\mathrm{col}$ being the collision cost and the last term being the distance of the optimized trajectory to the reinforcement learning algorithm's solution.

\begin{figure}[h!]
\vspace{2mm}
\centering
\def\svgwidth{8.0cm}
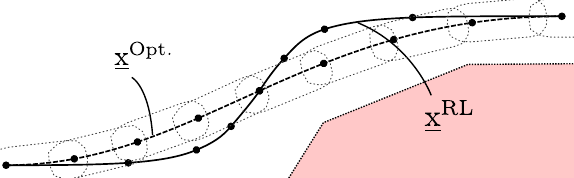
\caption{State-space trajectory provided by the reinforcement learning algorithm and the trajectory generatey by the post-optimization. }
\label{fig:optimization}
\end{figure}

Collisions with other vehicles, as well as with boundaries, are included in the collision term $\mathrm{\mathbf{J}}_\mathrm{col}$. The collision term uses the state history of all agents $[\underline{\mathrm{x}}^\mathrm{other, 1}, \dots, \underline{\mathrm{x}}^\mathrm{other, M}]$ that have been recorded stepping the environment during reinforcement learning.
Thus, the interaction with other vehicles is implicitly integrated into the local optimization method. Collision costs are imposed as soon as the ego-vehicle crosses a pre-defined safety-distance threshold.

The optimization outputs an optimal control input sequence $\underline{\mathrm{a}}^{*} = [\underline{\mathrm{a}}_0, \dots, \underline{\mathrm{a}}_\mathrm{N}]$  that generates the optimal state-space trajectory $\underline{\mathrm{x}}^{*} = [\underline{\mathrm{x}}_0, \dots, \underline{\mathrm{x}}_\mathrm{N}]$. After the post-optimization has finished an additional collision-check is performed to guarantee a high level of safety.

\section{Results and Evaluation}
In this section, we first present the used simulation framework and scenario chosen for training, applying and, evaluating the novel approach. Next, details of the used stochastic policy-based reinforcement learning algorithm -- the Soft Actor-Critic (SAC) -- are presented. Finally, we show and discuss the results of the novel approach that combines the reinforcement learning algorithm with the post-optimization.

\subsection{Simulation and Scenario}
A semantic simulation framework capable of simulating multiple agents has been developed and released in the course of this work \cite{Hart2019}. The framework is capable of simulating structured and unstructured environments and, thus, enables simulating a wide range of scenarios. Each agent in the environment is either controlled internally using a behavior model or can be controlled externally by providing system inputs. In this work, all agents but the ego-agent are controlled internally and drive with a constant velocity or behave according to the Intelligent Driver Model (IDM). The ego-vehicle is controlled externally and uses a single-track model that requires the steering angle and acceleration as system inputs ($\underline{\mathrm{a}} = [\delta, a]$). 

The initial states of all agents are uniformly sampled and, thus, the initial position, velocity, and angle in every episode are varied according to the specified range. In this work, we sample the speed in a range of $v \in [4m/s, 6m/s]$. The desired speed for all agents is set to $5m/s$. Furthermore, each agent has a pre-defined reference path that implicitly motivates each agent's goal.

The simulation framework is executed in an episodical fashion. An episode is counted as successful once the maximum number of steps has been reached without the ego-vehicle causing a collision with any other object. Furthermore, an observer assigns rewards to all events that occur during simulation and also calculates a concatenated state-space $\underline{\mathrm{s}}_t = [(x, y, v_{x}, v_{y})_0, \dots, (x, y, v_{x}, v_{y})_\mathrm{M}]$ with $\mathrm{M}$ being the number of vehicles. In this work, the concatenated state space includes the Cartesian coordinates $x,y$ as well as the velocities $v_x,v_y$ of each vehicle. The rewards are designed to avoid collisions, generate a comfortable driving behavior and to be close to the original reinforcement learning solution. To model the aforementioned characteristics, the reward values are set to

\begin{itemize}
\item $r_\mathrm{max, vel} = 10$ for the deviation to the desired velocity,
\item $r_\mathrm{max, ref} = 10$ for the deviation to the reference path,
\item $r_\mathrm{col} = -100$ for collisions with other objects and,
\item $r_\mathrm{jerk} = -0.1$ for the jerk of the trajectory.
\end{itemize}
Since $r_\mathrm{max, vel}$ and $r_\mathrm{max, ref}$ are the only positive rewards, the maximum achievable reward per epidose can be calculated using $r_\mathrm{total} = \sum_\mathrm{i=0}^{\mathrm{N}} r_\mathrm{max, vel} + r_\mathrm{max, ref}$ with $\mathrm{N}$ being the number of steps in a single episode. With the rewards presented above, the maximum achievable reward per episode with a hundred steps is $r_\mathrm{total} = 2000$. However, if there is an initial deviation to the reference path or to the desired velocity, the maximum achievable reward is smaller.

\subsection{Policy-based Reinforcement Learning}
In this work, we use the Soft Actor-Critic (SAC) policy-based reinforcement learning algorithm. It uses a distributional actor neural network having three layers (512, 512, 128) and a deterministic critic network also having three layers (256, 256, 128). The distributional network uses a normal distribution on the output layer and all other layers use the standard ReLU-activation function. In order to limit the input space for the dynamic model to feasible ranges, we limit the steering-angle $\delta \in [-0.2, 0.2]$ and the acceleration $a \in [-1, 1]$. To evaluate and quantify the results obtained by the SAC algorithm, we train and evaluate the following scenarios:

\begin{enumerate}
\item Two-lane merging having two other vehicles, 
\item two-lane merging having three other vehicles and,
\item highway merging having five other vehicles.
\end{enumerate}

\begin{figure}[h]
\centering
\resizebox {!} {6.5cm} {\input{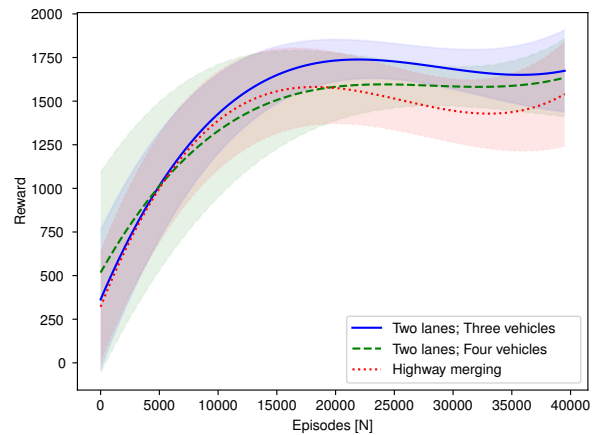}}
\caption{The average rewards obtained during training by the SAC algorithm.}
\label{fig:rewards}
\end{figure}

Figure \ref{fig:rewards}, shows the average reward obtained during training plotted over the episode numbers. It can be seen, that the reward increases for all of the above-presented scenarios and that the algorithm can handle all scenarios well.

Quantitative results of applying the SAC algorithm to the above-presented scenarios are shown in Table \ref{tab:table_1}. As expected the agent performs best, the less complex the scenario is. Despite this, the SAC algorithm is able to perform well in more complex scenarios, such as the highway merging scenario having many vehicles and multiple lanes.

\begin{table}[h]
\centering
\begin{tabular}{ c | c | c }
  Scenario & Success-rate [\%] & Avg. Reward \\
  \hline
  Two-lane merging (Three vehicles) & 99.99 \% & 1816.70 \\
  \hline
  Two-lane merging (Four vehicles) & 97.31 \% & 1827.36 \\
  \hline
  Highway merging & 96.05 \% & 1722.91 \\
\end{tabular}
\caption{For evaluation, the scenarios were run $1,000$ times.}
\label{tab:table_1}
\end{table}

In Figure \ref{fig:results}, a spatial plot of the highway scenario is shown with the ego-vehicle being controlled by the policy-based reinforcement learning algorithm (SAC) and the other agents using the Intelligent Driver Model (IDM) as behavior policy.

\begin{figure}[h]
\centering
\resizebox {!} {6.5cm} {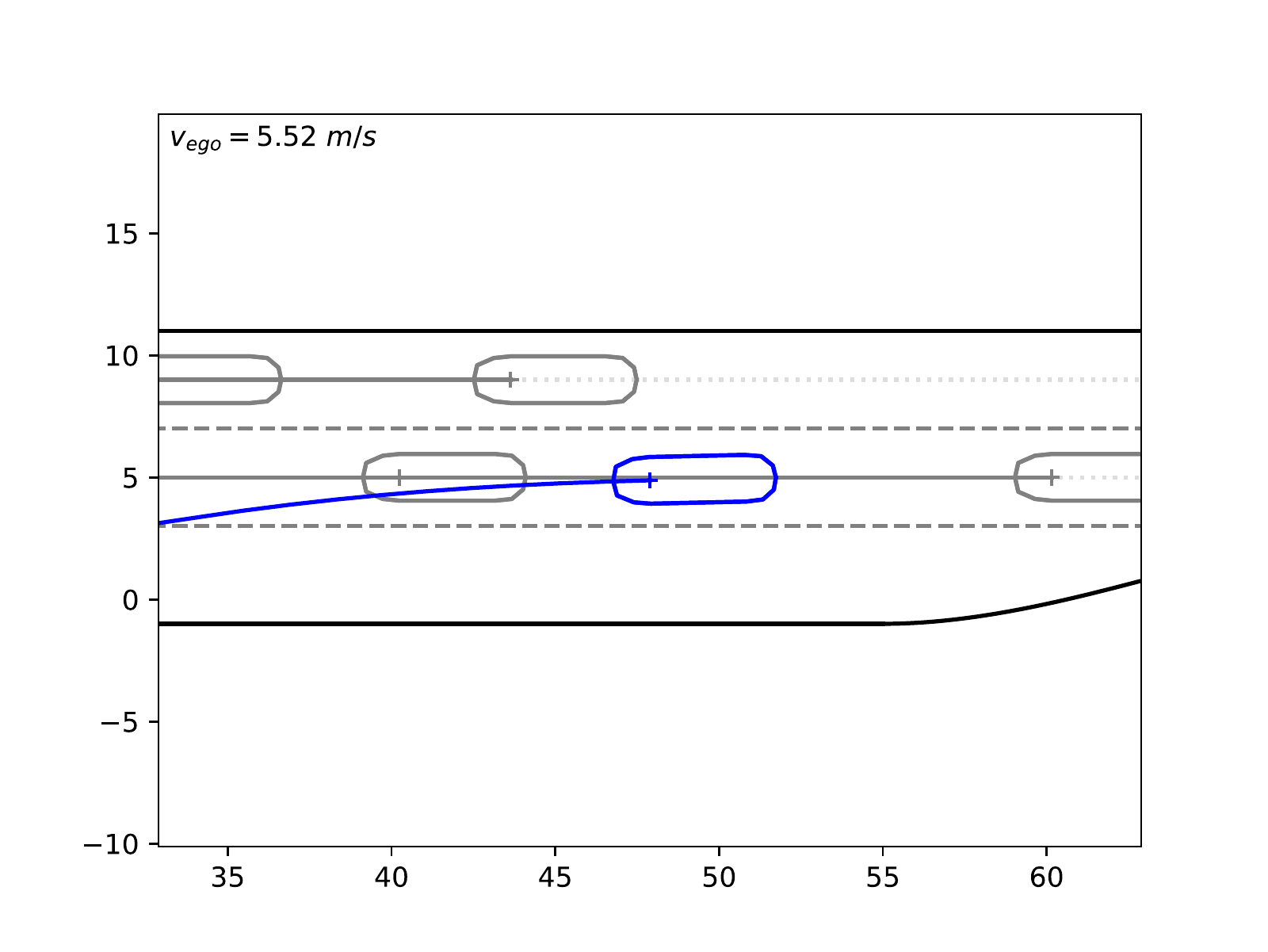}
\caption{Resulting behavior produced by the policy-based reinforcement learning approach is depicted in blue. All other vehicles behaving according to the IDM are shown in gray.}
\label{fig:results}
\end{figure}

The inference time executing the actor-network a single time on an i7-6700HQ processor having 2.60GHz and using Ubuntu 16.04 takes on average about $2.24ms$. The next sub-section will present the results of combining the policy-based reinforcement learning with the post-optimization method.

\subsection{Non-linear Post-Optimization}
The results of the novel approach combining policy-based reinforcement learning with a post-optimization are presented and discussed in this sub-section. We use two of the above-presented scenarios for the evaluation -- the two-lane with four vehicles and the highway scenario. The optimizer we use in this work is an unconstrained least-squares solver \cite{ceres-solver}. 

\begin{figure}[h]
\vspace{2mm}
\centering
\resizebox {!} {6.5cm} {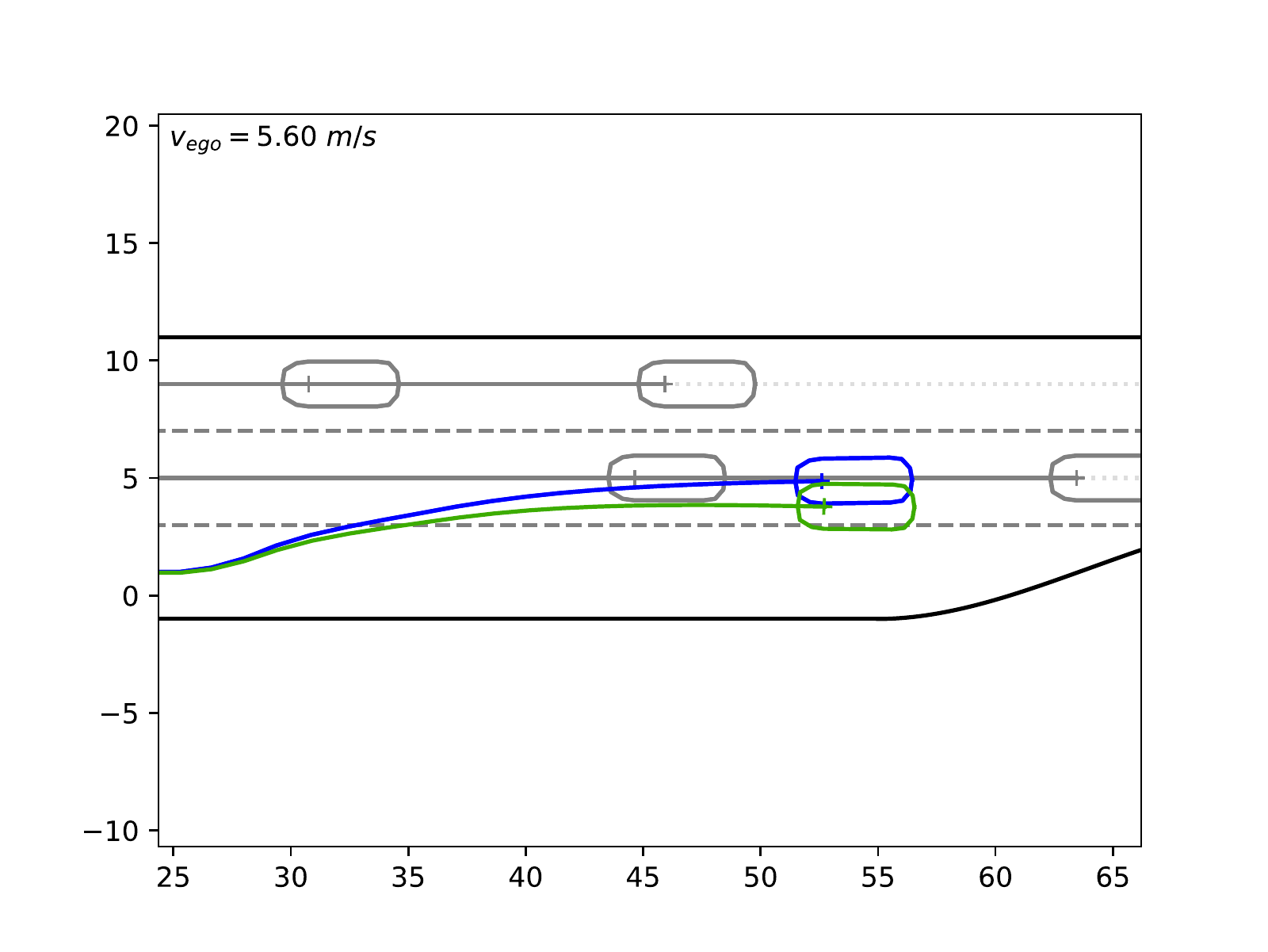}
\caption{The reinforcement learning algorithm's solution is depicted in blue and the optimizer's solution in green. The optimizer pushes the trajectory away from the other vehicle and additionally smoothens the trajectory.}
\label{fig:results}
\end{figure}
\begin{figure}[h]
\centering
\resizebox {!} {6.5cm} {\input{figures/opt_results.pgf}}
\caption{The control input sequences produced by the reinforcement learning algorithm and the optimization.}
\label{fig:results_velocity}
\end{figure}
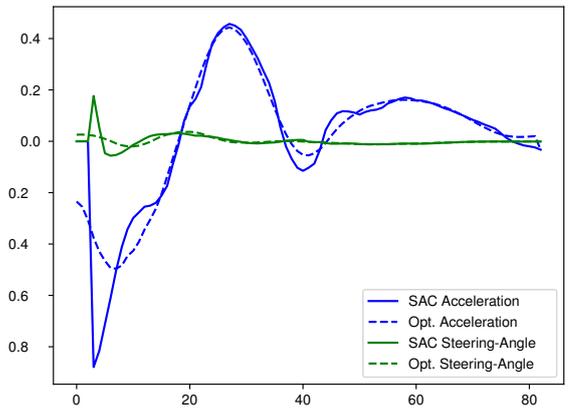

In Figure \ref{fig:results}, a spatial plot of the highway scenario is shown. The reinforcement learning algorithm's solution is depicted in blue and the optimized solution in green. Due to the additional safety-distance in the optimization, the trajectory is pushed further away from the other vehicle. The control input sequence produced by the learned policy of the reinforcement learning algorithm as well as the control input sequence of the optimization is shown in Figure \ref{fig:results_velocity}. Due to the additional jerk term in the objective function of the optimization, the resulting input sequences of the optimization are smoothened.

Quantitative results of the novel approach are shown in Table \ref{tab:table_2}. Due to the optimizer pushing the ego-vehicle away from other vehicles and boundaries, the collision rate drops to zero for the give time-horizon. Additionally, the jerk values are reduced and, thus, the comfort of the passengers is increased significantly. In case of a non-convergence of the optimizer, an emergency maneuver, such as emergency-braking could be performed. 

\begin{table}[h]
\centering
\begin{tabular}{ c | c | c }
  Scenario & Jerk. Opt./Jerk SAC [\%] & Jerk Opt. \\
  \hline
  Two-lane merging & 0.026 & 9.7 \\
  Highway merging & 0.034 & 7.75 \\
\end{tabular}
\caption{Results of the novel approach applied to the two-lane and highway scenario.}
\label{tab:table_2}
\end{table}

As shown in the previous sub-section, the inference-time of the policy-network is on average about $2.24ms$. Therefore, it would take around $112ms$ to produce a trajectory having $50$ time-steps. Contrary to this it takes the non-linear local optimizer on average about $206ms$ to solve a single homotopy class with $N = 40$ time-steps \cite{Hart2018}. Thus, choosing a homotopy class using reinforcement learning and obtaining an approximately optimal solution for the Markov decision process is more efficient than solving multiple homotopy classes. Furthermore, by using Graphics Processing Units (GPUs) the speed of both methods could be further improved -- during training as well as during application.

\section{Conclusion}
In this work, we showed that a stochastic policy-based reinforcement learning algorithm is able to solve complex scenarios, such as merging lanes with multiple other vehicles present. It learns implicitly and without any prior knowledge how to behave and which homotopy class to choose based on its collected experiences. In order to improve the solution provided by the reinforcement learning and to make it applicable in safety-critical applications, we used a non-linear post-optimization. By using the solution of the reinforcement learning algorithm in the optimization, a homotopy class is implicitly chosen. Thus, only a single objective has to be solved using local optimization instead of all in order to find the global optimum. Moreover, the comfort of the reinforcement learning algorithm's solution is improved due to the post-optimization. Furthermore, the optimizer's convergence can be evaluated for the use in safety-critical applications. 

In future research should be conducted on the generalization capabilities of the reinforcement learning to novel scnearios and to a varying number of vehicles. Moreover, occlusions and uncertainties could be modeled in the environment as well as more sophisticated behaviors of the other agents.

\printbibliography
\end{document}

%% file: figures/01_intro.pdf_tex
\begingroup%
  \makeatletter%
  \providecommand\color[2][]{%
    \errmessage{(Inkscape) Color is used for the text in Inkscape, but the package 'color.sty' is not loaded}%
    \renewcommand\color[2][]{}%
  }%
  \providecommand\transparent[1]{%
    \errmessage{(Inkscape) Transparency is used (non-zero) for the text in Inkscape, but the package 'transparent.sty' is not loaded}%
    \renewcommand\transparent[1]{}%
  }%
  \providecommand\rotatebox[2]{#2}%
  \ifx\svgwidth\undefined%
    \setlength{\unitlength}{426.79713452bp}%
    \ifx\svgscale\undefined%
      \relax%
    \else%
      \setlength{\unitlength}{\unitlength * \real{\svgscale}}%
    \fi%
  \else%
    \setlength{\unitlength}{\svgwidth}%
  \fi%
  \global\let\svgwidth\undefined%
  \global\let\svgscale\undefined%
  \makeatother%
  \begin{picture}(1,0.42829974)%
    \put(0,0){\includegraphics[width=\unitlength,page=1]{01_intro.pdf}}%
  \end{picture}%
\endgroup%

%% file: figures/opt.pdf_tex
\begingroup%
  \makeatletter%
  \providecommand\color[2][]{%
    \errmessage{(Inkscape) Color is used for the text in Inkscape, but the package 'color.sty' is not loaded}%
    \renewcommand\color[2][]{}%
  }%
  \providecommand\transparent[1]{%
    \errmessage{(Inkscape) Transparency is used (non-zero) for the text in Inkscape, but the package 'transparent.sty' is not loaded}%
    \renewcommand\transparent[1]{}%
  }%
  \providecommand\rotatebox[2]{#2}%
  \ifx\svgwidth\undefined%
    \setlength{\unitlength}{165.08571354bp}%
    \ifx\svgscale\undefined%
      \relax%
    \else%
      \setlength{\unitlength}{\unitlength * \real{\svgscale}}%
    \fi%
  \else%
    \setlength{\unitlength}{\svgwidth}%
  \fi%
  \global\let\svgwidth\undefined%
  \global\let\svgscale\undefined%
  \makeatother%
  \begin{picture}(1,0.30944962)%
    \put(0,0){\includegraphics[width=\unitlength,page=1]{opt.pdf}}%
  \end{picture}%
\endgroup%

%% file: figures/new_xy_res.pdf_tex
\begingroup%
  \makeatletter%
  \providecommand\color[2][]{%
    \errmessage{(Inkscape) Color is used for the text in Inkscape, but the package 'color.sty' is not loaded}%
    \renewcommand\color[2][]{}%
  }%
  \providecommand\transparent[1]{%
    \errmessage{(Inkscape) Transparency is used (non-zero) for the text in Inkscape, but the package 'transparent.sty' is not loaded}%
    \renewcommand\transparent[1]{}%
  }%
  \providecommand\rotatebox[2]{#2}%
  \ifx\svgwidth\undefined%
    \setlength{\unitlength}{460.79998779bp}%
    \ifx\svgscale\undefined%
      \relax%
    \else%
      \setlength{\unitlength}{\unitlength * \real{\svgscale}}%
    \fi%
  \else%
    \setlength{\unitlength}{\svgwidth}%
  \fi%
  \global\let\svgwidth\undefined%
  \global\let\svgscale\undefined%
  \makeatother%
  \begin{picture}(1,0.75000003)%
    \put(0,0){\includegraphics[width=\unitlength,page=1]{new_xy_res.pdf}}%
  \end{picture}%
\endgroup%

%% file: figures/new_opt_res.pdf_tex
\begingroup%
  \makeatletter%
  \providecommand\color[2][]{%
    \errmessage{(Inkscape) Color is used for the text in Inkscape, but the package 'color.sty' is not loaded}%
    \renewcommand\color[2][]{}%
  }%
  \providecommand\transparent[1]{%
    \errmessage{(Inkscape) Transparency is used (non-zero) for the text in Inkscape, but the package 'transparent.sty' is not loaded}%
    \renewcommand\transparent[1]{}%
  }%
  \providecommand\rotatebox[2]{#2}%
  \ifx\svgwidth\undefined%
    \setlength{\unitlength}{460.79998779bp}%
    \ifx\svgscale\undefined%
      \relax%
    \else%
      \setlength{\unitlength}{\unitlength * \real{\svgscale}}%
    \fi%
  \else%
    \setlength{\unitlength}{\svgwidth}%
  \fi%
  \global\let\svgwidth\undefined%
  \global\let\svgscale\undefined%
  \makeatother%
  \begin{picture}(1,0.75000003)%
    \put(0,0){\includegraphics[width=\unitlength,page=1]{new_opt_res.pdf}}%
  \end{picture}%
\endgroup%

%% file: figures/opt_results.pgf
\begingroup%
\makeatletter%
\begin{pgfpicture}%
\pgfpathrectangle{\pgfpointorigin}{\pgfqpoint{6.400000in}{4.800000in}}%
\pgfusepath{use as bounding box, clip}%
\begin{pgfscope}%
\pgfsetbuttcap%
\pgfsetmiterjoin%
\definecolor{currentfill}{rgb}{1.000000,1.000000,1.000000}%
\pgfsetfillcolor{currentfill}%
\pgfsetlinewidth{0.000000pt}%
\definecolor{currentstroke}{rgb}{1.000000,1.000000,1.000000}%
\pgfsetstrokecolor{currentstroke}%
\pgfsetdash{}{0pt}%
\pgfpathmoveto{\pgfqpoint{0.000000in}{0.000000in}}%
\pgfpathlineto{\pgfqpoint{6.400000in}{0.000000in}}%
\pgfpathlineto{\pgfqpoint{6.400000in}{4.800000in}}%
\pgfpathlineto{\pgfqpoint{0.000000in}{4.800000in}}%
\pgfpathclose%
\pgfusepath{fill}%
\end{pgfscope}%
\begin{pgfscope}%
\pgfsetbuttcap%
\pgfsetmiterjoin%
\definecolor{currentfill}{rgb}{1.000000,1.000000,1.000000}%
\pgfsetfillcolor{currentfill}%
\pgfsetlinewidth{0.000000pt}%
\definecolor{currentstroke}{rgb}{0.000000,0.000000,0.000000}%
\pgfsetstrokecolor{currentstroke}%
\pgfsetstrokeopacity{0.000000}%
\pgfsetdash{}{0pt}%
\pgfpathmoveto{\pgfqpoint{0.800000in}{0.528000in}}%
\pgfpathlineto{\pgfqpoint{5.760000in}{0.528000in}}%
\pgfpathlineto{\pgfqpoint{5.760000in}{4.224000in}}%
\pgfpathlineto{\pgfqpoint{0.800000in}{4.224000in}}%
\pgfpathclose%
\pgfusepath{fill}%
\end{pgfscope}%
\begin{pgfscope}%
\pgfsetbuttcap%
\pgfsetroundjoin%
\definecolor{currentfill}{rgb}{0.000000,0.000000,0.000000}%
\pgfsetfillcolor{currentfill}%
\pgfsetlinewidth{0.803000pt}%
\definecolor{currentstroke}{rgb}{0.000000,0.000000,0.000000}%
\pgfsetstrokecolor{currentstroke}%
\pgfsetdash{}{0pt}%
\pgfsys@defobject{currentmarker}{\pgfqpoint{0.000000in}{-0.048611in}}{\pgfqpoint{0.000000in}{0.000000in}}{%
\pgfpathmoveto{\pgfqpoint{0.000000in}{0.000000in}}%
\pgfpathlineto{\pgfqpoint{0.000000in}{-0.048611in}}%
\pgfusepath{stroke,fill}%
}%
\begin{pgfscope}%
\pgfsys@transformshift{1.025455in}{0.528000in}%
\pgfsys@useobject{currentmarker}{}%
\end{pgfscope}%
\end{pgfscope}%
\begin{pgfscope}%
\definecolor{textcolor}{rgb}{0.000000,0.000000,0.000000}%
\pgfsetstrokecolor{textcolor}%
\pgfsetfillcolor{textcolor}%
\pgftext[x=1.025455in,y=0.430778in,,top]{\color{textcolor}\sffamily\fontsize{10.000000}{12.000000}\selectfont 0}%
\end{pgfscope}%
\begin{pgfscope}%
\pgfsetbuttcap%
\pgfsetroundjoin%
\definecolor{currentfill}{rgb}{0.000000,0.000000,0.000000}%
\pgfsetfillcolor{currentfill}%
\pgfsetlinewidth{0.803000pt}%
\definecolor{currentstroke}{rgb}{0.000000,0.000000,0.000000}%
\pgfsetstrokecolor{currentstroke}%
\pgfsetdash{}{0pt}%
\pgfsys@defobject{currentmarker}{\pgfqpoint{0.000000in}{-0.048611in}}{\pgfqpoint{0.000000in}{0.000000in}}{%
\pgfpathmoveto{\pgfqpoint{0.000000in}{0.000000in}}%
\pgfpathlineto{\pgfqpoint{0.000000in}{-0.048611in}}%
\pgfusepath{stroke,fill}%
}%
\begin{pgfscope}%
\pgfsys@transformshift{2.125233in}{0.528000in}%
\pgfsys@useobject{currentmarker}{}%
\end{pgfscope}%
\end{pgfscope}%
\begin{pgfscope}%
\definecolor{textcolor}{rgb}{0.000000,0.000000,0.000000}%
\pgfsetstrokecolor{textcolor}%
\pgfsetfillcolor{textcolor}%
\pgftext[x=2.125233in,y=0.430778in,,top]{\color{textcolor}\sffamily\fontsize{10.000000}{12.000000}\selectfont 20}%
\end{pgfscope}%
\begin{pgfscope}%
\pgfsetbuttcap%
\pgfsetroundjoin%
\definecolor{currentfill}{rgb}{0.000000,0.000000,0.000000}%
\pgfsetfillcolor{currentfill}%
\pgfsetlinewidth{0.803000pt}%
\definecolor{currentstroke}{rgb}{0.000000,0.000000,0.000000}%
\pgfsetstrokecolor{currentstroke}%
\pgfsetdash{}{0pt}%
\pgfsys@defobject{currentmarker}{\pgfqpoint{0.000000in}{-0.048611in}}{\pgfqpoint{0.000000in}{0.000000in}}{%
\pgfpathmoveto{\pgfqpoint{0.000000in}{0.000000in}}%
\pgfpathlineto{\pgfqpoint{0.000000in}{-0.048611in}}%
\pgfusepath{stroke,fill}%
}%
\begin{pgfscope}%
\pgfsys@transformshift{3.225011in}{0.528000in}%
\pgfsys@useobject{currentmarker}{}%
\end{pgfscope}%
\end{pgfscope}%
\begin{pgfscope}%
\definecolor{textcolor}{rgb}{0.000000,0.000000,0.000000}%
\pgfsetstrokecolor{textcolor}%
\pgfsetfillcolor{textcolor}%
\pgftext[x=3.225011in,y=0.430778in,,top]{\color{textcolor}\sffamily\fontsize{10.000000}{12.000000}\selectfont 40}%
\end{pgfscope}%
\begin{pgfscope}%
\pgfsetbuttcap%
\pgfsetroundjoin%
\definecolor{currentfill}{rgb}{0.000000,0.000000,0.000000}%
\pgfsetfillcolor{currentfill}%
\pgfsetlinewidth{0.803000pt}%
\definecolor{currentstroke}{rgb}{0.000000,0.000000,0.000000}%
\pgfsetstrokecolor{currentstroke}%
\pgfsetdash{}{0pt}%
\pgfsys@defobject{currentmarker}{\pgfqpoint{0.000000in}{-0.048611in}}{\pgfqpoint{0.000000in}{0.000000in}}{%
\pgfpathmoveto{\pgfqpoint{0.000000in}{0.000000in}}%
\pgfpathlineto{\pgfqpoint{0.000000in}{-0.048611in}}%
\pgfusepath{stroke,fill}%
}%
\begin{pgfscope}%
\pgfsys@transformshift{4.324789in}{0.528000in}%
\pgfsys@useobject{currentmarker}{}%
\end{pgfscope}%
\end{pgfscope}%
\begin{pgfscope}%
\definecolor{textcolor}{rgb}{0.000000,0.000000,0.000000}%
\pgfsetstrokecolor{textcolor}%
\pgfsetfillcolor{textcolor}%
\pgftext[x=4.324789in,y=0.430778in,,top]{\color{textcolor}\sffamily\fontsize{10.000000}{12.000000}\selectfont 60}%
\end{pgfscope}%
\begin{pgfscope}%
\pgfsetbuttcap%
\pgfsetroundjoin%
\definecolor{currentfill}{rgb}{0.000000,0.000000,0.000000}%
\pgfsetfillcolor{currentfill}%
\pgfsetlinewidth{0.803000pt}%
\definecolor{currentstroke}{rgb}{0.000000,0.000000,0.000000}%
\pgfsetstrokecolor{currentstroke}%
\pgfsetdash{}{0pt}%
\pgfsys@defobject{currentmarker}{\pgfqpoint{0.000000in}{-0.048611in}}{\pgfqpoint{0.000000in}{0.000000in}}{%
\pgfpathmoveto{\pgfqpoint{0.000000in}{0.000000in}}%
\pgfpathlineto{\pgfqpoint{0.000000in}{-0.048611in}}%
\pgfusepath{stroke,fill}%
}%
\begin{pgfscope}%
\pgfsys@transformshift{5.424568in}{0.528000in}%
\pgfsys@useobject{currentmarker}{}%
\end{pgfscope}%
\end{pgfscope}%
\begin{pgfscope}%
\definecolor{textcolor}{rgb}{0.000000,0.000000,0.000000}%
\pgfsetstrokecolor{textcolor}%
\pgfsetfillcolor{textcolor}%
\pgftext[x=5.424568in,y=0.430778in,,top]{\color{textcolor}\sffamily\fontsize{10.000000}{12.000000}\selectfont 80}%
\end{pgfscope}%
\begin{pgfscope}%
\pgfsetbuttcap%
\pgfsetroundjoin%
\definecolor{currentfill}{rgb}{0.000000,0.000000,0.000000}%
\pgfsetfillcolor{currentfill}%
\pgfsetlinewidth{0.803000pt}%
\definecolor{currentstroke}{rgb}{0.000000,0.000000,0.000000}%
\pgfsetstrokecolor{currentstroke}%
\pgfsetdash{}{0pt}%
\pgfsys@defobject{currentmarker}{\pgfqpoint{-0.048611in}{0.000000in}}{\pgfqpoint{0.000000in}{0.000000in}}{%
\pgfpathmoveto{\pgfqpoint{0.000000in}{0.000000in}}%
\pgfpathlineto{\pgfqpoint{-0.048611in}{0.000000in}}%
\pgfusepath{stroke,fill}%
}%
\begin{pgfscope}%
\pgfsys@transformshift{0.800000in}{0.894656in}%
\pgfsys@useobject{currentmarker}{}%
\end{pgfscope}%
\end{pgfscope}%
\begin{pgfscope}%
\definecolor{textcolor}{rgb}{0.000000,0.000000,0.000000}%
\pgfsetstrokecolor{textcolor}%
\pgfsetfillcolor{textcolor}%
\pgftext[x=0.365525in,y=0.841895in,left,base]{\color{textcolor}\sffamily\fontsize{10.000000}{12.000000}\selectfont −0.8}%
\end{pgfscope}%
\begin{pgfscope}%
\pgfsetbuttcap%
\pgfsetroundjoin%
\definecolor{currentfill}{rgb}{0.000000,0.000000,0.000000}%
\pgfsetfillcolor{currentfill}%
\pgfsetlinewidth{0.803000pt}%
\definecolor{currentstroke}{rgb}{0.000000,0.000000,0.000000}%
\pgfsetstrokecolor{currentstroke}%
\pgfsetdash{}{0pt}%
\pgfsys@defobject{currentmarker}{\pgfqpoint{-0.048611in}{0.000000in}}{\pgfqpoint{0.000000in}{0.000000in}}{%
\pgfpathmoveto{\pgfqpoint{0.000000in}{0.000000in}}%
\pgfpathlineto{\pgfqpoint{-0.048611in}{0.000000in}}%
\pgfusepath{stroke,fill}%
}%
\begin{pgfscope}%
\pgfsys@transformshift{0.800000in}{1.397598in}%
\pgfsys@useobject{currentmarker}{}%
\end{pgfscope}%
\end{pgfscope}%
\begin{pgfscope}%
\definecolor{textcolor}{rgb}{0.000000,0.000000,0.000000}%
\pgfsetstrokecolor{textcolor}%
\pgfsetfillcolor{textcolor}%
\pgftext[x=0.365525in,y=1.344836in,left,base]{\color{textcolor}\sffamily\fontsize{10.000000}{12.000000}\selectfont −0.6}%
\end{pgfscope}%
\begin{pgfscope}%
\pgfsetbuttcap%
\pgfsetroundjoin%
\definecolor{currentfill}{rgb}{0.000000,0.000000,0.000000}%
\pgfsetfillcolor{currentfill}%
\pgfsetlinewidth{0.803000pt}%
\definecolor{currentstroke}{rgb}{0.000000,0.000000,0.000000}%
\pgfsetstrokecolor{currentstroke}%
\pgfsetdash{}{0pt}%
\pgfsys@defobject{currentmarker}{\pgfqpoint{-0.048611in}{0.000000in}}{\pgfqpoint{0.000000in}{0.000000in}}{%
\pgfpathmoveto{\pgfqpoint{0.000000in}{0.000000in}}%
\pgfpathlineto{\pgfqpoint{-0.048611in}{0.000000in}}%
\pgfusepath{stroke,fill}%
}%
\begin{pgfscope}%
\pgfsys@transformshift{0.800000in}{1.900540in}%
\pgfsys@useobject{currentmarker}{}%
\end{pgfscope}%
\end{pgfscope}%
\begin{pgfscope}%
\definecolor{textcolor}{rgb}{0.000000,0.000000,0.000000}%
\pgfsetstrokecolor{textcolor}%
\pgfsetfillcolor{textcolor}%
\pgftext[x=0.365525in,y=1.847778in,left,base]{\color{textcolor}\sffamily\fontsize{10.000000}{12.000000}\selectfont −0.4}%
\end{pgfscope}%
\begin{pgfscope}%
\pgfsetbuttcap%
\pgfsetroundjoin%
\definecolor{currentfill}{rgb}{0.000000,0.000000,0.000000}%
\pgfsetfillcolor{currentfill}%
\pgfsetlinewidth{0.803000pt}%
\definecolor{currentstroke}{rgb}{0.000000,0.000000,0.000000}%
\pgfsetstrokecolor{currentstroke}%
\pgfsetdash{}{0pt}%
\pgfsys@defobject{currentmarker}{\pgfqpoint{-0.048611in}{0.000000in}}{\pgfqpoint{0.000000in}{0.000000in}}{%
\pgfpathmoveto{\pgfqpoint{0.000000in}{0.000000in}}%
\pgfpathlineto{\pgfqpoint{-0.048611in}{0.000000in}}%
\pgfusepath{stroke,fill}%
}%
\begin{pgfscope}%
\pgfsys@transformshift{0.800000in}{2.403482in}%
\pgfsys@useobject{currentmarker}{}%
\end{pgfscope}%
\end{pgfscope}%
\begin{pgfscope}%
\definecolor{textcolor}{rgb}{0.000000,0.000000,0.000000}%
\pgfsetstrokecolor{textcolor}%
\pgfsetfillcolor{textcolor}%
\pgftext[x=0.365525in,y=2.350720in,left,base]{\color{textcolor}\sffamily\fontsize{10.000000}{12.000000}\selectfont −0.2}%
\end{pgfscope}%
\begin{pgfscope}%
\pgfsetbuttcap%
\pgfsetroundjoin%
\definecolor{currentfill}{rgb}{0.000000,0.000000,0.000000}%
\pgfsetfillcolor{currentfill}%
\pgfsetlinewidth{0.803000pt}%
\definecolor{currentstroke}{rgb}{0.000000,0.000000,0.000000}%
\pgfsetstrokecolor{currentstroke}%
\pgfsetdash{}{0pt}%
\pgfsys@defobject{currentmarker}{\pgfqpoint{-0.048611in}{0.000000in}}{\pgfqpoint{0.000000in}{0.000000in}}{%
\pgfpathmoveto{\pgfqpoint{0.000000in}{0.000000in}}%
\pgfpathlineto{\pgfqpoint{-0.048611in}{0.000000in}}%
\pgfusepath{stroke,fill}%
}%
\begin{pgfscope}%
\pgfsys@transformshift{0.800000in}{2.906424in}%
\pgfsys@useobject{currentmarker}{}%
\end{pgfscope}%
\end{pgfscope}%
\begin{pgfscope}%
\definecolor{textcolor}{rgb}{0.000000,0.000000,0.000000}%
\pgfsetstrokecolor{textcolor}%
\pgfsetfillcolor{textcolor}%
\pgftext[x=0.481898in,y=2.853662in,left,base]{\color{textcolor}\sffamily\fontsize{10.000000}{12.000000}\selectfont 0.0}%
\end{pgfscope}%
\begin{pgfscope}%
\pgfsetbuttcap%
\pgfsetroundjoin%
\definecolor{currentfill}{rgb}{0.000000,0.000000,0.000000}%
\pgfsetfillcolor{currentfill}%
\pgfsetlinewidth{0.803000pt}%
\definecolor{currentstroke}{rgb}{0.000000,0.000000,0.000000}%
\pgfsetstrokecolor{currentstroke}%
\pgfsetdash{}{0pt}%
\pgfsys@defobject{currentmarker}{\pgfqpoint{-0.048611in}{0.000000in}}{\pgfqpoint{0.000000in}{0.000000in}}{%
\pgfpathmoveto{\pgfqpoint{0.000000in}{0.000000in}}%
\pgfpathlineto{\pgfqpoint{-0.048611in}{0.000000in}}%
\pgfusepath{stroke,fill}%
}%
\begin{pgfscope}%
\pgfsys@transformshift{0.800000in}{3.409365in}%
\pgfsys@useobject{currentmarker}{}%
\end{pgfscope}%
\end{pgfscope}%
\begin{pgfscope}%
\definecolor{textcolor}{rgb}{0.000000,0.000000,0.000000}%
\pgfsetstrokecolor{textcolor}%
\pgfsetfillcolor{textcolor}%
\pgftext[x=0.481898in,y=3.356604in,left,base]{\color{textcolor}\sffamily\fontsize{10.000000}{12.000000}\selectfont 0.2}%
\end{pgfscope}%
\begin{pgfscope}%
\pgfsetbuttcap%
\pgfsetroundjoin%
\definecolor{currentfill}{rgb}{0.000000,0.000000,0.000000}%
\pgfsetfillcolor{currentfill}%
\pgfsetlinewidth{0.803000pt}%
\definecolor{currentstroke}{rgb}{0.000000,0.000000,0.000000}%
\pgfsetstrokecolor{currentstroke}%
\pgfsetdash{}{0pt}%
\pgfsys@defobject{currentmarker}{\pgfqpoint{-0.048611in}{0.000000in}}{\pgfqpoint{0.000000in}{0.000000in}}{%
\pgfpathmoveto{\pgfqpoint{0.000000in}{0.000000in}}%
\pgfpathlineto{\pgfqpoint{-0.048611in}{0.000000in}}%
\pgfusepath{stroke,fill}%
}%
\begin{pgfscope}%
\pgfsys@transformshift{0.800000in}{3.912307in}%
\pgfsys@useobject{currentmarker}{}%
\end{pgfscope}%
\end{pgfscope}%
\begin{pgfscope}%
\definecolor{textcolor}{rgb}{0.000000,0.000000,0.000000}%
\pgfsetstrokecolor{textcolor}%
\pgfsetfillcolor{textcolor}%
\pgftext[x=0.481898in,y=3.859546in,left,base]{\color{textcolor}\sffamily\fontsize{10.000000}{12.000000}\selectfont 0.4}%
\end{pgfscope}%
\begin{pgfscope}%
\pgfpathrectangle{\pgfqpoint{0.800000in}{0.528000in}}{\pgfqpoint{4.960000in}{3.696000in}}%
\pgfusepath{clip}%
\pgfsetrectcap%
\pgfsetroundjoin%
\pgfsetlinewidth{1.505625pt}%
\definecolor{currentstroke}{rgb}{0.000000,0.000000,1.000000}%
\pgfsetstrokecolor{currentstroke}%
\pgfsetdash{}{0pt}%
\pgfpathmoveto{\pgfqpoint{1.025455in}{2.906424in}}%
\pgfpathlineto{\pgfqpoint{1.080443in}{2.906424in}}%
\pgfpathlineto{\pgfqpoint{1.135432in}{2.906424in}}%
\pgfpathlineto{\pgfqpoint{1.190421in}{0.696000in}}%
\pgfpathlineto{\pgfqpoint{1.245410in}{0.854494in}}%
\pgfpathlineto{\pgfqpoint{1.300399in}{1.115049in}}%
\pgfpathlineto{\pgfqpoint{1.355388in}{1.373377in}}%
\pgfpathlineto{\pgfqpoint{1.410377in}{1.647104in}}%
\pgfpathlineto{\pgfqpoint{1.465366in}{1.873963in}}%
\pgfpathlineto{\pgfqpoint{1.520355in}{2.046588in}}%
\pgfpathlineto{\pgfqpoint{1.575344in}{2.155253in}}%
\pgfpathlineto{\pgfqpoint{1.630333in}{2.209764in}}%
\pgfpathlineto{\pgfqpoint{1.685322in}{2.266010in}}%
\pgfpathlineto{\pgfqpoint{1.740310in}{2.276178in}}%
\pgfpathlineto{\pgfqpoint{1.795299in}{2.303880in}}%
\pgfpathlineto{\pgfqpoint{1.850288in}{2.373405in}}%
\pgfpathlineto{\pgfqpoint{1.905277in}{2.469380in}}%
\pgfpathlineto{\pgfqpoint{1.960266in}{2.681033in}}%
\pgfpathlineto{\pgfqpoint{2.015255in}{2.871205in}}%
\pgfpathlineto{\pgfqpoint{2.070244in}{3.106561in}}%
\pgfpathlineto{\pgfqpoint{2.125233in}{3.247932in}}%
\pgfpathlineto{\pgfqpoint{2.180222in}{3.319028in}}%
\pgfpathlineto{\pgfqpoint{2.235211in}{3.436037in}}%
\pgfpathlineto{\pgfqpoint{2.290200in}{3.671139in}}%
\pgfpathlineto{\pgfqpoint{2.345188in}{3.870782in}}%
\pgfpathlineto{\pgfqpoint{2.400177in}{3.966253in}}%
\pgfpathlineto{\pgfqpoint{2.455166in}{4.023796in}}%
\pgfpathlineto{\pgfqpoint{2.510155in}{4.056000in}}%
\pgfpathlineto{\pgfqpoint{2.565144in}{4.037350in}}%
\pgfpathlineto{\pgfqpoint{2.620133in}{3.998614in}}%
\pgfpathlineto{\pgfqpoint{2.675122in}{3.916795in}}%
\pgfpathlineto{\pgfqpoint{2.730111in}{3.818357in}}%
\pgfpathlineto{\pgfqpoint{2.785100in}{3.718202in}}%
\pgfpathlineto{\pgfqpoint{2.840089in}{3.594798in}}%
\pgfpathlineto{\pgfqpoint{2.895078in}{3.479519in}}%
\pgfpathlineto{\pgfqpoint{2.950067in}{3.300580in}}%
\pgfpathlineto{\pgfqpoint{3.005055in}{3.044652in}}%
\pgfpathlineto{\pgfqpoint{3.060044in}{2.854440in}}%
\pgfpathlineto{\pgfqpoint{3.115033in}{2.734180in}}%
\pgfpathlineto{\pgfqpoint{3.170022in}{2.647587in}}%
\pgfpathlineto{\pgfqpoint{3.225011in}{2.616784in}}%
\pgfpathlineto{\pgfqpoint{3.280000in}{2.647770in}}%
\pgfpathlineto{\pgfqpoint{3.334989in}{2.687541in}}%
\pgfpathlineto{\pgfqpoint{3.389978in}{2.824289in}}%
\pgfpathlineto{\pgfqpoint{3.444967in}{3.016235in}}%
\pgfpathlineto{\pgfqpoint{3.499956in}{3.113428in}}%
\pgfpathlineto{\pgfqpoint{3.554945in}{3.180280in}}%
\pgfpathlineto{\pgfqpoint{3.609933in}{3.200959in}}%
\pgfpathlineto{\pgfqpoint{3.664922in}{3.198603in}}%
\pgfpathlineto{\pgfqpoint{3.719911in}{3.188360in}}%
\pgfpathlineto{\pgfqpoint{3.774900in}{3.167910in}}%
\pgfpathlineto{\pgfqpoint{3.829889in}{3.193341in}}%
\pgfpathlineto{\pgfqpoint{3.884878in}{3.208537in}}%
\pgfpathlineto{\pgfqpoint{3.939867in}{3.216267in}}%
\pgfpathlineto{\pgfqpoint{3.994856in}{3.235388in}}%
\pgfpathlineto{\pgfqpoint{4.049845in}{3.277023in}}%
\pgfpathlineto{\pgfqpoint{4.104834in}{3.296731in}}%
\pgfpathlineto{\pgfqpoint{4.159823in}{3.320959in}}%
\pgfpathlineto{\pgfqpoint{4.214812in}{3.334488in}}%
\pgfpathlineto{\pgfqpoint{4.269800in}{3.327252in}}%
\pgfpathlineto{\pgfqpoint{4.324789in}{3.306848in}}%
\pgfpathlineto{\pgfqpoint{4.379778in}{3.305322in}}%
\pgfpathlineto{\pgfqpoint{4.434767in}{3.291850in}}%
\pgfpathlineto{\pgfqpoint{4.489756in}{3.275354in}}%
\pgfpathlineto{\pgfqpoint{4.544745in}{3.252521in}}%
\pgfpathlineto{\pgfqpoint{4.599734in}{3.229287in}}%
\pgfpathlineto{\pgfqpoint{4.654723in}{3.212158in}}%
\pgfpathlineto{\pgfqpoint{4.709712in}{3.188304in}}%
\pgfpathlineto{\pgfqpoint{4.764701in}{3.165666in}}%
\pgfpathlineto{\pgfqpoint{4.819690in}{3.140810in}}%
\pgfpathlineto{\pgfqpoint{4.874678in}{3.116115in}}%
\pgfpathlineto{\pgfqpoint{4.929667in}{3.096068in}}%
\pgfpathlineto{\pgfqpoint{4.984656in}{3.071503in}}%
\pgfpathlineto{\pgfqpoint{5.039645in}{3.051221in}}%
\pgfpathlineto{\pgfqpoint{5.094634in}{3.029161in}}%
\pgfpathlineto{\pgfqpoint{5.149623in}{2.982968in}}%
\pgfpathlineto{\pgfqpoint{5.204612in}{2.937910in}}%
\pgfpathlineto{\pgfqpoint{5.259601in}{2.905048in}}%
\pgfpathlineto{\pgfqpoint{5.314590in}{2.884152in}}%
\pgfpathlineto{\pgfqpoint{5.369579in}{2.866410in}}%
\pgfpathlineto{\pgfqpoint{5.424568in}{2.856818in}}%
\pgfpathlineto{\pgfqpoint{5.479557in}{2.846178in}}%
\pgfpathlineto{\pgfqpoint{5.534545in}{2.824064in}}%
\pgfusepath{stroke}%
\end{pgfscope}%
\begin{pgfscope}%
\pgfpathrectangle{\pgfqpoint{0.800000in}{0.528000in}}{\pgfqpoint{4.960000in}{3.696000in}}%
\pgfusepath{clip}%
\pgfsetbuttcap%
\pgfsetroundjoin%
\pgfsetlinewidth{1.505625pt}%
\definecolor{currentstroke}{rgb}{0.000000,0.000000,1.000000}%
\pgfsetstrokecolor{currentstroke}%
\pgfsetdash{{5.550000pt}{2.400000pt}}{0.000000pt}%
\pgfpathmoveto{\pgfqpoint{1.025455in}{2.316677in}}%
\pgfpathlineto{\pgfqpoint{1.080443in}{2.264329in}}%
\pgfpathlineto{\pgfqpoint{1.135432in}{2.134250in}}%
\pgfpathlineto{\pgfqpoint{1.190421in}{1.962500in}}%
\pgfpathlineto{\pgfqpoint{1.245410in}{1.827445in}}%
\pgfpathlineto{\pgfqpoint{1.300399in}{1.736176in}}%
\pgfpathlineto{\pgfqpoint{1.355388in}{1.665657in}}%
\pgfpathlineto{\pgfqpoint{1.410377in}{1.659172in}}%
\pgfpathlineto{\pgfqpoint{1.465366in}{1.690772in}}%
\pgfpathlineto{\pgfqpoint{1.520355in}{1.784016in}}%
\pgfpathlineto{\pgfqpoint{1.575344in}{1.834197in}}%
\pgfpathlineto{\pgfqpoint{1.630333in}{1.921870in}}%
\pgfpathlineto{\pgfqpoint{1.685322in}{2.041232in}}%
\pgfpathlineto{\pgfqpoint{1.740310in}{2.139101in}}%
\pgfpathlineto{\pgfqpoint{1.795299in}{2.245605in}}%
\pgfpathlineto{\pgfqpoint{1.850288in}{2.388467in}}%
\pgfpathlineto{\pgfqpoint{1.905277in}{2.561295in}}%
\pgfpathlineto{\pgfqpoint{1.960266in}{2.738738in}}%
\pgfpathlineto{\pgfqpoint{2.015255in}{2.913685in}}%
\pgfpathlineto{\pgfqpoint{2.070244in}{3.085683in}}%
\pgfpathlineto{\pgfqpoint{2.125233in}{3.258365in}}%
\pgfpathlineto{\pgfqpoint{2.180222in}{3.428557in}}%
\pgfpathlineto{\pgfqpoint{2.235211in}{3.590840in}}%
\pgfpathlineto{\pgfqpoint{2.290200in}{3.733897in}}%
\pgfpathlineto{\pgfqpoint{2.345188in}{3.855726in}}%
\pgfpathlineto{\pgfqpoint{2.400177in}{3.947643in}}%
\pgfpathlineto{\pgfqpoint{2.455166in}{4.003855in}}%
\pgfpathlineto{\pgfqpoint{2.510155in}{4.021401in}}%
\pgfpathlineto{\pgfqpoint{2.565144in}{4.000781in}}%
\pgfpathlineto{\pgfqpoint{2.620133in}{3.947172in}}%
\pgfpathlineto{\pgfqpoint{2.675122in}{3.869361in}}%
\pgfpathlineto{\pgfqpoint{2.730111in}{3.766133in}}%
\pgfpathlineto{\pgfqpoint{2.785100in}{3.642130in}}%
\pgfpathlineto{\pgfqpoint{2.840089in}{3.507250in}}%
\pgfpathlineto{\pgfqpoint{2.895078in}{3.366623in}}%
\pgfpathlineto{\pgfqpoint{2.950067in}{3.224637in}}%
\pgfpathlineto{\pgfqpoint{3.005055in}{3.092568in}}%
\pgfpathlineto{\pgfqpoint{3.060044in}{2.975068in}}%
\pgfpathlineto{\pgfqpoint{3.115033in}{2.878394in}}%
\pgfpathlineto{\pgfqpoint{3.170022in}{2.810697in}}%
\pgfpathlineto{\pgfqpoint{3.225011in}{2.774103in}}%
\pgfpathlineto{\pgfqpoint{3.280000in}{2.768313in}}%
\pgfpathlineto{\pgfqpoint{3.334989in}{2.789037in}}%
\pgfpathlineto{\pgfqpoint{3.389978in}{2.829851in}}%
\pgfpathlineto{\pgfqpoint{3.444967in}{2.889412in}}%
\pgfpathlineto{\pgfqpoint{3.499956in}{2.960588in}}%
\pgfpathlineto{\pgfqpoint{3.554945in}{3.027170in}}%
\pgfpathlineto{\pgfqpoint{3.609933in}{3.081781in}}%
\pgfpathlineto{\pgfqpoint{3.664922in}{3.127558in}}%
\pgfpathlineto{\pgfqpoint{3.719911in}{3.167835in}}%
\pgfpathlineto{\pgfqpoint{3.774900in}{3.204587in}}%
\pgfpathlineto{\pgfqpoint{3.829889in}{3.234632in}}%
\pgfpathlineto{\pgfqpoint{3.884878in}{3.255176in}}%
\pgfpathlineto{\pgfqpoint{3.939867in}{3.270045in}}%
\pgfpathlineto{\pgfqpoint{3.994856in}{3.284765in}}%
\pgfpathlineto{\pgfqpoint{4.049845in}{3.297118in}}%
\pgfpathlineto{\pgfqpoint{4.104834in}{3.306847in}}%
\pgfpathlineto{\pgfqpoint{4.159823in}{3.310966in}}%
\pgfpathlineto{\pgfqpoint{4.214812in}{3.312025in}}%
\pgfpathlineto{\pgfqpoint{4.269800in}{3.310036in}}%
\pgfpathlineto{\pgfqpoint{4.324789in}{3.305275in}}%
\pgfpathlineto{\pgfqpoint{4.379778in}{3.298070in}}%
\pgfpathlineto{\pgfqpoint{4.434767in}{3.289351in}}%
\pgfpathlineto{\pgfqpoint{4.489756in}{3.277176in}}%
\pgfpathlineto{\pgfqpoint{4.544745in}{3.260885in}}%
\pgfpathlineto{\pgfqpoint{4.599734in}{3.241055in}}%
\pgfpathlineto{\pgfqpoint{4.654723in}{3.218518in}}%
\pgfpathlineto{\pgfqpoint{4.709712in}{3.193726in}}%
\pgfpathlineto{\pgfqpoint{4.764701in}{3.166418in}}%
\pgfpathlineto{\pgfqpoint{4.819690in}{3.136483in}}%
\pgfpathlineto{\pgfqpoint{4.874678in}{3.104575in}}%
\pgfpathlineto{\pgfqpoint{4.929667in}{3.073417in}}%
\pgfpathlineto{\pgfqpoint{4.984656in}{3.044234in}}%
\pgfpathlineto{\pgfqpoint{5.039645in}{3.017339in}}%
\pgfpathlineto{\pgfqpoint{5.094634in}{2.992913in}}%
\pgfpathlineto{\pgfqpoint{5.149623in}{2.973408in}}%
\pgfpathlineto{\pgfqpoint{5.204612in}{2.959829in}}%
\pgfpathlineto{\pgfqpoint{5.259601in}{2.952141in}}%
\pgfpathlineto{\pgfqpoint{5.314590in}{2.948979in}}%
\pgfpathlineto{\pgfqpoint{5.369579in}{2.949499in}}%
\pgfpathlineto{\pgfqpoint{5.424568in}{2.952598in}}%
\pgfpathlineto{\pgfqpoint{5.479557in}{2.956232in}}%
\pgfpathlineto{\pgfqpoint{5.534545in}{2.824064in}}%
\pgfusepath{stroke}%
\end{pgfscope}%
\begin{pgfscope}%
\pgfpathrectangle{\pgfqpoint{0.800000in}{0.528000in}}{\pgfqpoint{4.960000in}{3.696000in}}%
\pgfusepath{clip}%
\pgfsetrectcap%
\pgfsetroundjoin%
\pgfsetlinewidth{1.505625pt}%
\definecolor{currentstroke}{rgb}{0.000000,0.501961,0.000000}%
\pgfsetstrokecolor{currentstroke}%
\pgfsetdash{}{0pt}%
\pgfpathmoveto{\pgfqpoint{1.025455in}{2.906424in}}%
\pgfpathlineto{\pgfqpoint{1.080443in}{2.906424in}}%
\pgfpathlineto{\pgfqpoint{1.135432in}{2.906424in}}%
\pgfpathlineto{\pgfqpoint{1.190421in}{3.349981in}}%
\pgfpathlineto{\pgfqpoint{1.245410in}{3.042034in}}%
\pgfpathlineto{\pgfqpoint{1.300399in}{2.789343in}}%
\pgfpathlineto{\pgfqpoint{1.355388in}{2.764946in}}%
\pgfpathlineto{\pgfqpoint{1.410377in}{2.772816in}}%
\pgfpathlineto{\pgfqpoint{1.465366in}{2.796838in}}%
\pgfpathlineto{\pgfqpoint{1.520355in}{2.832284in}}%
\pgfpathlineto{\pgfqpoint{1.575344in}{2.871187in}}%
\pgfpathlineto{\pgfqpoint{1.630333in}{2.899792in}}%
\pgfpathlineto{\pgfqpoint{1.685322in}{2.931314in}}%
\pgfpathlineto{\pgfqpoint{1.740310in}{2.957374in}}%
\pgfpathlineto{\pgfqpoint{1.795299in}{2.970164in}}%
\pgfpathlineto{\pgfqpoint{1.850288in}{2.974904in}}%
\pgfpathlineto{\pgfqpoint{1.905277in}{2.976493in}}%
\pgfpathlineto{\pgfqpoint{1.960266in}{2.982270in}}%
\pgfpathlineto{\pgfqpoint{2.015255in}{2.985277in}}%
\pgfpathlineto{\pgfqpoint{2.070244in}{2.980176in}}%
\pgfpathlineto{\pgfqpoint{2.125233in}{2.971920in}}%
\pgfpathlineto{\pgfqpoint{2.180222in}{2.961346in}}%
\pgfpathlineto{\pgfqpoint{2.235211in}{2.961449in}}%
\pgfpathlineto{\pgfqpoint{2.290200in}{2.954075in}}%
\pgfpathlineto{\pgfqpoint{2.345188in}{2.949536in}}%
\pgfpathlineto{\pgfqpoint{2.400177in}{2.940572in}}%
\pgfpathlineto{\pgfqpoint{2.455166in}{2.927231in}}%
\pgfpathlineto{\pgfqpoint{2.510155in}{2.919423in}}%
\pgfpathlineto{\pgfqpoint{2.565144in}{2.908813in}}%
\pgfpathlineto{\pgfqpoint{2.620133in}{2.901107in}}%
\pgfpathlineto{\pgfqpoint{2.675122in}{2.891826in}}%
\pgfpathlineto{\pgfqpoint{2.730111in}{2.887401in}}%
\pgfpathlineto{\pgfqpoint{2.785100in}{2.885173in}}%
\pgfpathlineto{\pgfqpoint{2.840089in}{2.888784in}}%
\pgfpathlineto{\pgfqpoint{2.895078in}{2.892297in}}%
\pgfpathlineto{\pgfqpoint{2.950067in}{2.895226in}}%
\pgfpathlineto{\pgfqpoint{3.005055in}{2.902820in}}%
\pgfpathlineto{\pgfqpoint{3.060044in}{2.911842in}}%
\pgfpathlineto{\pgfqpoint{3.115033in}{2.915565in}}%
\pgfpathlineto{\pgfqpoint{3.170022in}{2.920023in}}%
\pgfpathlineto{\pgfqpoint{3.225011in}{2.919667in}}%
\pgfpathlineto{\pgfqpoint{3.280000in}{2.901162in}}%
\pgfpathlineto{\pgfqpoint{3.334989in}{2.896259in}}%
\pgfpathlineto{\pgfqpoint{3.389978in}{2.899405in}}%
\pgfpathlineto{\pgfqpoint{3.444967in}{2.889259in}}%
\pgfpathlineto{\pgfqpoint{3.499956in}{2.882469in}}%
\pgfpathlineto{\pgfqpoint{3.554945in}{2.882433in}}%
\pgfpathlineto{\pgfqpoint{3.609933in}{2.884540in}}%
\pgfpathlineto{\pgfqpoint{3.664922in}{2.885407in}}%
\pgfpathlineto{\pgfqpoint{3.719911in}{2.883821in}}%
\pgfpathlineto{\pgfqpoint{3.774900in}{2.877784in}}%
\pgfpathlineto{\pgfqpoint{3.829889in}{2.878930in}}%
\pgfpathlineto{\pgfqpoint{3.884878in}{2.877653in}}%
\pgfpathlineto{\pgfqpoint{3.939867in}{2.878406in}}%
\pgfpathlineto{\pgfqpoint{3.994856in}{2.878200in}}%
\pgfpathlineto{\pgfqpoint{4.049845in}{2.876930in}}%
\pgfpathlineto{\pgfqpoint{4.104834in}{2.879181in}}%
\pgfpathlineto{\pgfqpoint{4.159823in}{2.883013in}}%
\pgfpathlineto{\pgfqpoint{4.214812in}{2.884052in}}%
\pgfpathlineto{\pgfqpoint{4.269800in}{2.884174in}}%
\pgfpathlineto{\pgfqpoint{4.324789in}{2.884034in}}%
\pgfpathlineto{\pgfqpoint{4.379778in}{2.882681in}}%
\pgfpathlineto{\pgfqpoint{4.434767in}{2.884155in}}%
\pgfpathlineto{\pgfqpoint{4.489756in}{2.886582in}}%
\pgfpathlineto{\pgfqpoint{4.544745in}{2.889240in}}%
\pgfpathlineto{\pgfqpoint{4.599734in}{2.891955in}}%
\pgfpathlineto{\pgfqpoint{4.654723in}{2.894252in}}%
\pgfpathlineto{\pgfqpoint{4.709712in}{2.896573in}}%
\pgfpathlineto{\pgfqpoint{4.764701in}{2.898646in}}%
\pgfpathlineto{\pgfqpoint{4.819690in}{2.900854in}}%
\pgfpathlineto{\pgfqpoint{4.874678in}{2.902114in}}%
\pgfpathlineto{\pgfqpoint{4.929667in}{2.902966in}}%
\pgfpathlineto{\pgfqpoint{4.984656in}{2.903813in}}%
\pgfpathlineto{\pgfqpoint{5.039645in}{2.904226in}}%
\pgfpathlineto{\pgfqpoint{5.094634in}{2.904684in}}%
\pgfpathlineto{\pgfqpoint{5.149623in}{2.904220in}}%
\pgfpathlineto{\pgfqpoint{5.204612in}{2.903800in}}%
\pgfpathlineto{\pgfqpoint{5.259601in}{2.903549in}}%
\pgfpathlineto{\pgfqpoint{5.314590in}{2.903161in}}%
\pgfpathlineto{\pgfqpoint{5.369579in}{2.903514in}}%
\pgfpathlineto{\pgfqpoint{5.424568in}{2.904149in}}%
\pgfpathlineto{\pgfqpoint{5.479557in}{2.903992in}}%
\pgfpathlineto{\pgfqpoint{5.534545in}{2.904181in}}%
\pgfusepath{stroke}%
\end{pgfscope}%
\begin{pgfscope}%
\pgfpathrectangle{\pgfqpoint{0.800000in}{0.528000in}}{\pgfqpoint{4.960000in}{3.696000in}}%
\pgfusepath{clip}%
\pgfsetbuttcap%
\pgfsetroundjoin%
\pgfsetlinewidth{1.505625pt}%
\definecolor{currentstroke}{rgb}{0.000000,0.501961,0.000000}%
\pgfsetstrokecolor{currentstroke}%
\pgfsetdash{{5.550000pt}{2.400000pt}}{0.000000pt}%
\pgfpathmoveto{\pgfqpoint{1.025455in}{2.970816in}}%
\pgfpathlineto{\pgfqpoint{1.080443in}{2.972143in}}%
\pgfpathlineto{\pgfqpoint{1.135432in}{2.966978in}}%
\pgfpathlineto{\pgfqpoint{1.190421in}{2.961978in}}%
\pgfpathlineto{\pgfqpoint{1.245410in}{2.947497in}}%
\pgfpathlineto{\pgfqpoint{1.300399in}{2.930187in}}%
\pgfpathlineto{\pgfqpoint{1.355388in}{2.909251in}}%
\pgfpathlineto{\pgfqpoint{1.410377in}{2.884758in}}%
\pgfpathlineto{\pgfqpoint{1.465366in}{2.866531in}}%
\pgfpathlineto{\pgfqpoint{1.520355in}{2.857173in}}%
\pgfpathlineto{\pgfqpoint{1.575344in}{2.856335in}}%
\pgfpathlineto{\pgfqpoint{1.630333in}{2.863668in}}%
\pgfpathlineto{\pgfqpoint{1.685322in}{2.880154in}}%
\pgfpathlineto{\pgfqpoint{1.740310in}{2.900696in}}%
\pgfpathlineto{\pgfqpoint{1.795299in}{2.928415in}}%
\pgfpathlineto{\pgfqpoint{1.850288in}{2.955458in}}%
\pgfpathlineto{\pgfqpoint{1.905277in}{2.971094in}}%
\pgfpathlineto{\pgfqpoint{1.960266in}{2.983798in}}%
\pgfpathlineto{\pgfqpoint{2.015255in}{2.995149in}}%
\pgfpathlineto{\pgfqpoint{2.070244in}{2.997075in}}%
\pgfpathlineto{\pgfqpoint{2.125233in}{2.998932in}}%
\pgfpathlineto{\pgfqpoint{2.180222in}{2.995214in}}%
\pgfpathlineto{\pgfqpoint{2.235211in}{2.980794in}}%
\pgfpathlineto{\pgfqpoint{2.290200in}{2.962676in}}%
\pgfpathlineto{\pgfqpoint{2.345188in}{2.943288in}}%
\pgfpathlineto{\pgfqpoint{2.400177in}{2.928643in}}%
\pgfpathlineto{\pgfqpoint{2.455166in}{2.914001in}}%
\pgfpathlineto{\pgfqpoint{2.510155in}{2.901362in}}%
\pgfpathlineto{\pgfqpoint{2.565144in}{2.895771in}}%
\pgfpathlineto{\pgfqpoint{2.620133in}{2.894205in}}%
\pgfpathlineto{\pgfqpoint{2.675122in}{2.892852in}}%
\pgfpathlineto{\pgfqpoint{2.730111in}{2.892898in}}%
\pgfpathlineto{\pgfqpoint{2.785100in}{2.895443in}}%
\pgfpathlineto{\pgfqpoint{2.840089in}{2.898187in}}%
\pgfpathlineto{\pgfqpoint{2.895078in}{2.902371in}}%
\pgfpathlineto{\pgfqpoint{2.950067in}{2.905092in}}%
\pgfpathlineto{\pgfqpoint{3.005055in}{2.906935in}}%
\pgfpathlineto{\pgfqpoint{3.060044in}{2.908820in}}%
\pgfpathlineto{\pgfqpoint{3.115033in}{2.908405in}}%
\pgfpathlineto{\pgfqpoint{3.170022in}{2.906347in}}%
\pgfpathlineto{\pgfqpoint{3.225011in}{2.903997in}}%
\pgfpathlineto{\pgfqpoint{3.280000in}{2.901107in}}%
\pgfpathlineto{\pgfqpoint{3.334989in}{2.899035in}}%
\pgfpathlineto{\pgfqpoint{3.389978in}{2.896675in}}%
\pgfpathlineto{\pgfqpoint{3.444967in}{2.893420in}}%
\pgfpathlineto{\pgfqpoint{3.499956in}{2.891101in}}%
\pgfpathlineto{\pgfqpoint{3.554945in}{2.888813in}}%
\pgfpathlineto{\pgfqpoint{3.609933in}{2.885987in}}%
\pgfpathlineto{\pgfqpoint{3.664922in}{2.883938in}}%
\pgfpathlineto{\pgfqpoint{3.719911in}{2.880514in}}%
\pgfpathlineto{\pgfqpoint{3.774900in}{2.879291in}}%
\pgfpathlineto{\pgfqpoint{3.829889in}{2.876520in}}%
\pgfpathlineto{\pgfqpoint{3.884878in}{2.876321in}}%
\pgfpathlineto{\pgfqpoint{3.939867in}{2.876025in}}%
\pgfpathlineto{\pgfqpoint{3.994856in}{2.877818in}}%
\pgfpathlineto{\pgfqpoint{4.049845in}{2.878610in}}%
\pgfpathlineto{\pgfqpoint{4.104834in}{2.880035in}}%
\pgfpathlineto{\pgfqpoint{4.159823in}{2.881620in}}%
\pgfpathlineto{\pgfqpoint{4.214812in}{2.882219in}}%
\pgfpathlineto{\pgfqpoint{4.269800in}{2.882281in}}%
\pgfpathlineto{\pgfqpoint{4.324789in}{2.882366in}}%
\pgfpathlineto{\pgfqpoint{4.379778in}{2.886207in}}%
\pgfpathlineto{\pgfqpoint{4.434767in}{2.887499in}}%
\pgfpathlineto{\pgfqpoint{4.489756in}{2.890057in}}%
\pgfpathlineto{\pgfqpoint{4.544745in}{2.892091in}}%
\pgfpathlineto{\pgfqpoint{4.599734in}{2.892624in}}%
\pgfpathlineto{\pgfqpoint{4.654723in}{2.894076in}}%
\pgfpathlineto{\pgfqpoint{4.709712in}{2.895918in}}%
\pgfpathlineto{\pgfqpoint{4.764701in}{2.897730in}}%
\pgfpathlineto{\pgfqpoint{4.819690in}{2.899259in}}%
\pgfpathlineto{\pgfqpoint{4.874678in}{2.901307in}}%
\pgfpathlineto{\pgfqpoint{4.929667in}{2.902411in}}%
\pgfpathlineto{\pgfqpoint{4.984656in}{2.902248in}}%
\pgfpathlineto{\pgfqpoint{5.039645in}{2.902672in}}%
\pgfpathlineto{\pgfqpoint{5.094634in}{2.903129in}}%
\pgfpathlineto{\pgfqpoint{5.149623in}{2.903179in}}%
\pgfpathlineto{\pgfqpoint{5.204612in}{2.903672in}}%
\pgfpathlineto{\pgfqpoint{5.259601in}{2.903976in}}%
\pgfpathlineto{\pgfqpoint{5.314590in}{2.904312in}}%
\pgfpathlineto{\pgfqpoint{5.369579in}{2.904564in}}%
\pgfpathlineto{\pgfqpoint{5.424568in}{2.904427in}}%
\pgfpathlineto{\pgfqpoint{5.479557in}{2.904295in}}%
\pgfpathlineto{\pgfqpoint{5.534545in}{2.904181in}}%
\pgfusepath{stroke}%
\end{pgfscope}%
\begin{pgfscope}%
\pgfsetrectcap%
\pgfsetmiterjoin%
\pgfsetlinewidth{0.803000pt}%
\definecolor{currentstroke}{rgb}{0.000000,0.000000,0.000000}%
\pgfsetstrokecolor{currentstroke}%
\pgfsetdash{}{0pt}%
\pgfpathmoveto{\pgfqpoint{0.800000in}{0.528000in}}%
\pgfpathlineto{\pgfqpoint{0.800000in}{4.224000in}}%
\pgfusepath{stroke}%
\end{pgfscope}%
\begin{pgfscope}%
\pgfsetrectcap%
\pgfsetmiterjoin%
\pgfsetlinewidth{0.803000pt}%
\definecolor{currentstroke}{rgb}{0.000000,0.000000,0.000000}%
\pgfsetstrokecolor{currentstroke}%
\pgfsetdash{}{0pt}%
\pgfpathmoveto{\pgfqpoint{5.760000in}{0.528000in}}%
\pgfpathlineto{\pgfqpoint{5.760000in}{4.224000in}}%
\pgfusepath{stroke}%
\end{pgfscope}%
\begin{pgfscope}%
\pgfsetrectcap%
\pgfsetmiterjoin%
\pgfsetlinewidth{0.803000pt}%
\definecolor{currentstroke}{rgb}{0.000000,0.000000,0.000000}%
\pgfsetstrokecolor{currentstroke}%
\pgfsetdash{}{0pt}%
\pgfpathmoveto{\pgfqpoint{0.800000in}{0.528000in}}%
\pgfpathlineto{\pgfqpoint{5.760000in}{0.528000in}}%
\pgfusepath{stroke}%
\end{pgfscope}%
\begin{pgfscope}%
\pgfsetrectcap%
\pgfsetmiterjoin%
\pgfsetlinewidth{0.803000pt}%
\definecolor{currentstroke}{rgb}{0.000000,0.000000,0.000000}%
\pgfsetstrokecolor{currentstroke}%
\pgfsetdash{}{0pt}%
\pgfpathmoveto{\pgfqpoint{0.800000in}{4.224000in}}%
\pgfpathlineto{\pgfqpoint{5.760000in}{4.224000in}}%
\pgfusepath{stroke}%
\end{pgfscope}%
\begin{pgfscope}%
\pgfsetbuttcap%
\pgfsetmiterjoin%
\definecolor{currentfill}{rgb}{1.000000,1.000000,1.000000}%
\pgfsetfillcolor{currentfill}%
\pgfsetfillopacity{0.800000}%
\pgfsetlinewidth{1.003750pt}%
\definecolor{currentstroke}{rgb}{0.800000,0.800000,0.800000}%
\pgfsetstrokecolor{currentstroke}%
\pgfsetstrokeopacity{0.800000}%
\pgfsetdash{}{0pt}%
\pgfpathmoveto{\pgfqpoint{3.835684in}{0.597444in}}%
\pgfpathlineto{\pgfqpoint{5.662778in}{0.597444in}}%
\pgfpathquadraticcurveto{\pgfqpoint{5.690556in}{0.597444in}}{\pgfqpoint{5.690556in}{0.625222in}}%
\pgfpathlineto{\pgfqpoint{5.690556in}{1.426762in}}%
\pgfpathquadraticcurveto{\pgfqpoint{5.690556in}{1.454540in}}{\pgfqpoint{5.662778in}{1.454540in}}%
\pgfpathlineto{\pgfqpoint{3.835684in}{1.454540in}}%
\pgfpathquadraticcurveto{\pgfqpoint{3.807906in}{1.454540in}}{\pgfqpoint{3.807906in}{1.426762in}}%
\pgfpathlineto{\pgfqpoint{3.807906in}{0.625222in}}%
\pgfpathquadraticcurveto{\pgfqpoint{3.807906in}{0.597444in}}{\pgfqpoint{3.835684in}{0.597444in}}%
\pgfpathclose%
\pgfusepath{stroke,fill}%
\end{pgfscope}%
\begin{pgfscope}%
\pgfsetrectcap%
\pgfsetroundjoin%
\pgfsetlinewidth{1.505625pt}%
\definecolor{currentstroke}{rgb}{0.000000,0.000000,1.000000}%
\pgfsetstrokecolor{currentstroke}%
\pgfsetdash{}{0pt}%
\pgfpathmoveto{\pgfqpoint{3.863461in}{1.342073in}}%
\pgfpathlineto{\pgfqpoint{4.141239in}{1.342073in}}%
\pgfusepath{stroke}%
\end{pgfscope}%
\begin{pgfscope}%
\definecolor{textcolor}{rgb}{0.000000,0.000000,0.000000}%
\pgfsetstrokecolor{textcolor}%
\pgfsetfillcolor{textcolor}%
\pgftext[x=4.252350in,y=1.293461in,left,base]{\color{textcolor}\sffamily\fontsize{10.000000}{12.000000}\selectfont SAC Acceleration}%
\end{pgfscope}%
\begin{pgfscope}%
\pgfsetbuttcap%
\pgfsetroundjoin%
\pgfsetlinewidth{1.505625pt}%
\definecolor{currentstroke}{rgb}{0.000000,0.000000,1.000000}%
\pgfsetstrokecolor{currentstroke}%
\pgfsetdash{{5.550000pt}{2.400000pt}}{0.000000pt}%
\pgfpathmoveto{\pgfqpoint{3.863461in}{1.138215in}}%
\pgfpathlineto{\pgfqpoint{4.141239in}{1.138215in}}%
\pgfusepath{stroke}%
\end{pgfscope}%
\begin{pgfscope}%
\definecolor{textcolor}{rgb}{0.000000,0.000000,0.000000}%
\pgfsetstrokecolor{textcolor}%
\pgfsetfillcolor{textcolor}%
\pgftext[x=4.252350in,y=1.089604in,left,base]{\color{textcolor}\sffamily\fontsize{10.000000}{12.000000}\selectfont Opt. Acceleration}%
\end{pgfscope}%
\begin{pgfscope}%
\pgfsetrectcap%
\pgfsetroundjoin%
\pgfsetlinewidth{1.505625pt}%
\definecolor{currentstroke}{rgb}{0.000000,0.501961,0.000000}%
\pgfsetstrokecolor{currentstroke}%
\pgfsetdash{}{0pt}%
\pgfpathmoveto{\pgfqpoint{3.863461in}{0.934358in}}%
\pgfpathlineto{\pgfqpoint{4.141239in}{0.934358in}}%
\pgfusepath{stroke}%
\end{pgfscope}%
\begin{pgfscope}%
\definecolor{textcolor}{rgb}{0.000000,0.000000,0.000000}%
\pgfsetstrokecolor{textcolor}%
\pgfsetfillcolor{textcolor}%
\pgftext[x=4.252350in,y=0.885747in,left,base]{\color{textcolor}\sffamily\fontsize{10.000000}{12.000000}\selectfont SAC Steering-Angle}%
\end{pgfscope}%
\begin{pgfscope}%
\pgfsetbuttcap%
\pgfsetroundjoin%
\pgfsetlinewidth{1.505625pt}%
\definecolor{currentstroke}{rgb}{0.000000,0.501961,0.000000}%
\pgfsetstrokecolor{currentstroke}%
\pgfsetdash{{5.550000pt}{2.400000pt}}{0.000000pt}%
\pgfpathmoveto{\pgfqpoint{3.863461in}{0.730501in}}%
\pgfpathlineto{\pgfqpoint{4.141239in}{0.730501in}}%
\pgfusepath{stroke}%
\end{pgfscope}%
\begin{pgfscope}%
\definecolor{textcolor}{rgb}{0.000000,0.000000,0.000000}%
\pgfsetstrokecolor{textcolor}%
\pgfsetfillcolor{textcolor}%
\pgftext[x=4.252350in,y=0.681890in,left,base]{\color{textcolor}\sffamily\fontsize{10.000000}{12.000000}\selectfont Opt. Steering-Angle}%
\end{pgfscope}%
\end{pgfpicture}%
\makeatother%
\endgroup%